\documentclass[10pt,twocolumn,letterpaper]{article}
\usepackage{lineno}
\usepackage{titlesec}
\usepackage{cvpr}
\definecolor{cvprblue}{rgb}{0.21,0.49,0.74}
\usepackage[pagebackref,breaklinks,colorlinks,allcolors=cvprblue]{hyperref}
\usepackage{bm}
\usepackage{multicol}
\usepackage{multirow}

\def\paperID{10602}
\def\confName{CVPR}
\def\confYear{2025}

\begin{document}
\title{FRESA: Feedforward Reconstruction of Personalized Skinned Avatars \\ from Few Images}

\author{Rong Wang$^{1,2 *}$ \and Fabian Prada$^{2}$ \and Ziyan Wang$^{2}$ \and Zhongshi Jiang$^{2}$ \and Chengxiang Yin$^{2}$ \and Junxuan Li$^{2}$ \and
Shunsuke Saito$^{2}$\and Igor Santesteban$^{2}$\and Javier Romero$^{2}$\and Rohan Joshi$^{2}$\and Hongdong Li$^{1}$\and Jason Saragih$^{2}$\and Yaser Sheikh$^{2}$ \and \vspace{0.3cm}
$^1$Australian National University \hspace{0.4cm} $^2$Meta Reality Labs Research}

\twocolumn[{%
\renewcommand\twocolumn[1][]{#1}%
\maketitle
\begin{center}
    \vspace{-1cm}
    \centering
    \captionsetup{type=figure}
    \includegraphics[width=\textwidth]{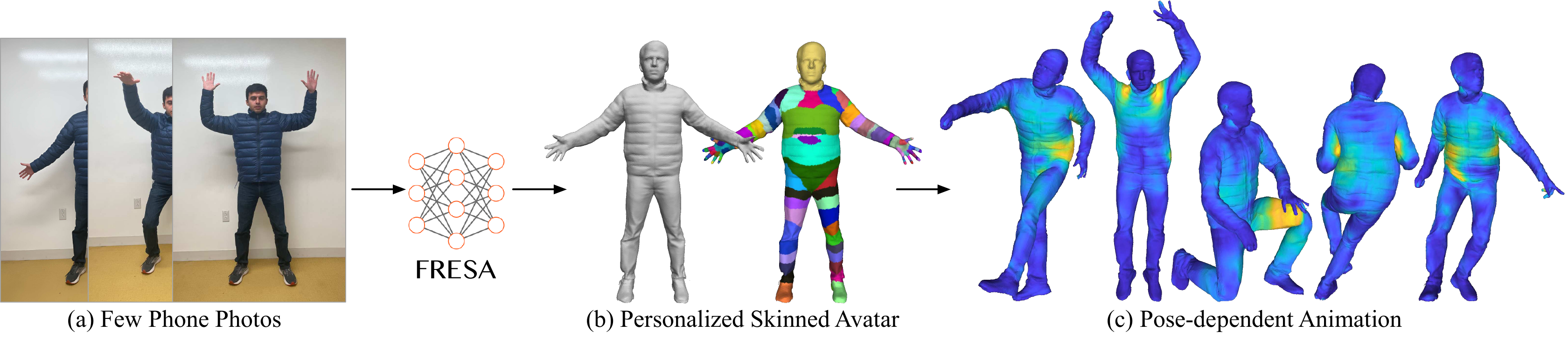}
    \captionof{figure}{\textbf{FRESA. }We present a novel method to reconstruct personalized skinned avatars with realistic pose-dependent animation all in a \emph{feed-forward} approach, which generalizes to causally taken phone photos without any fine-tuning. We visualize the predicted skinning weights associated with the most important joints in (b) and colormaps of per-vertex displacement magnitudes\protect\footnotemark during animation in (c).}
\end{center}%
}]
\maketitle

\begin{abstract}
We present a novel method for reconstructing personalized 3D human avatars with realistic animation from only a few images. Due to the large variations in body shapes, poses, and cloth types, existing methods mostly require hours of per-subject optimization during inference, which limits their practical applications. In contrast, we learn a universal prior from over a thousand clothed humans to achieve instant feedforward generation and zero-shot generalization. Specifically, instead of rigging the avatar with shared skinning weights, we jointly infer personalized avatar shape, skinning weights, and pose-dependent deformations, which effectively improves overall geometric fidelity and reduces deformation artifacts. Moreover, to normalize pose variations and resolve coupled ambiguity between canonical shapes and skinning weights, we design a 3D canonicalization process to produce pixel-aligned initial conditions, which helps to reconstruct fine-grained geometric details. We then propose a multi-frame feature aggregation to robustly reduce artifacts introduced in canonicalization and fuse a plausible avatar preserving person-specific identities. Finally, we train the model in an end-to-end framework on a large-scale capture dataset, which contains diverse human subjects paired with high-quality 3D scans. Extensive experiments show that our method generates more authentic reconstruction and animation than state-of-the-arts, and can be directly generalized to inputs from casually taken phone photos. Project page and code is available at \url{https://github.com/rongakowang/FRESA}.
\end{abstract}

\renewcommand{\thefootnote}{*}
\footnotetext{Work done while Rong Wang was an intern at Reality Labs Research.}
\renewcommand{\thefootnote}{\arabic{footnote}}
\footnotetext{Scales normalized across all vertices to highlight large deformation.}    
\section{Introduction}
\label{sec:intro}

3D digitization for clothed humans is highly demanded in many vision and graphics applications, such as extended reality (XR), virtual try-on, and telepresence \cite{lawrence2022state}. Since humans vary largely in body shapes and cloth types, the reconstructed avatars need to faithfully preserve these personalized details, and can be realistically animated robust to the underlying topology. To this end, traditional pipelines often require laborious artist designs \cite{collet2015high, 10.1007/978-3-031-72933-1_3} or controlled environments \cite{guo2019relightables, parger2021unoc}, which limits their use. In this work, we aim to automatically reconstruct animatable avatars from easily accessible sources, \emph{e.g.} casually taken phone photos, to better facilitate downstream applications.

Existing works in 3D clothed human reconstructions \cite{xiu2022icon,xiu2023econ,saito2019pifu,saito2020pifuhd, zhang2024global, zhang2024sifu} have demonstrated impressive quality and generalizability. While they mostly focus on reconstructing a static frame for a given pose, \cite{huang2020arch, he2021arch++, liao2023high} further extend to produce avatars more suitable for animation. This is achieved by reconstructing the avatar in a \emph{canonical} space associated with a rigged body template, and then animating it by target joint motions via the linear blend skinning (LBS) \cite{magnenat1988joint} method. However, all these works skin the avatar by querying the skinning weights from nearest template body vertices, which inevitably produces deformation artifacts in challenging poses and extreme body shapes. Recently, \cite{saito2021scanimate, chen2021snarf, xue2023nsf, wang2022arah} propose to jointly optimize personalized canonical shapes and skinning weights from posed 3D scans or images. However, due to the lack of a unified prior from different body shapes and cloth types \cite{wang2021metaavatar}, they are restricted to a per-subject setting and require hours of test-time optimization, which is computationally expensive.

In this paper, we present a novel method to reconstruct personalized skinned avatars for realistic animation in an efficient \emph{feed-forward} approach. The key to our method is a \emph{universal} clothed human model, which is learned from a large collection of human subjects while capable of jointly inferring personalized canonical avatar shapes, skinning weights, and pose-dependent deformation in animation. Specifically, to normalize pose variations in input images and ensure model generalizability, we first develop a 3D canonicalization process to produce pixel-aligned initial conditions for geometry and semantics references, which helps to reconstruct canonical avatars with fine-grained geometric details. While the canonicalization can produce artifacts due to noisy unposing, we propose to robustly reduce the artifacts and generate a plausible avatar by aggregating from multi-frame references. In this way, the fused feature can preserve information intrinsic to the person, \emph{e.g.} body shapes and cloth types, to better represent the person identity. Finally, to resolve the coupled ambiguity between the canonical geometry and skinning weights, \emph{i.e.} to avoid generating incorrect components that can be accidentally warped to a correct posed shape \cite{lin2022learning}, we adopt a multi-stage training process to jointly supervise the model with posed-space ground truths and canonical-space regularization. Leveraging the rich prior from the universal model, we achieve feed-forward reconstruction of avatars and animation without requiring expensive per-subject optimization.

Our key contributions can be summarized as follows. (\emph{i}) We present a novel method \textbf{FRESA}, which enables instant feed-forward reconstruction of 3D human avatars with personalized canonical shapes, skinning weights, and pose-dependent animation via a universal clothed human model, while achieving zero-shot generalization to inputs of only a few casually taken photos. (\emph{ii}) We propose a novel pipeline with explicit canonicalization and multi-frame aggregation, which robustly improves the fidelity and realism of resulting avatars. (\emph{iii}) We develop a large-scale clothed human dataset with diverse subjects and high-quality 3D scans to facilitate learning an effective universal prior. Extensive experiments show that our method outperforms state-of-the-arts in both reconstruction fidelity and animation quality.

\section{Related Works}
\textbf{3D Clothed Human Reconstruction. }Early works of 3D human reconstruction often utilize parametric human models \cite{loper2023smpl, pavlakos2019expressive} as a prior for the underlying body shapes. For instance, \cite{ma2020learning, bhatnagar2019multi} adopt a SMPL+D model to express cloth geometry as per-vertex displacements of the SMPL \cite{loper2023smpl} body template, while \cite{lahner2018deepwrinkles, alldieck2019tex2shape} use UV maps to further improve the mesh resolutions. However, these approaches assume a fixed topology and can not model loose clothes with large shape variation to the body template. In contrast, \cite{saito2019pifu, saito2020pifuhd, xiu2022icon, xiu2023econ, alldieck2022photorealistic} propose to leverage neural implicit fields \cite{park2019deepsdf, mescheder2019occupancy} for continuous surface representation, which can flexibly model diverse body shapes and clothes types. Such representation also supports fast mesh extraction via Marching Cubes \cite{lorensen1998marching} to enable explicit geometry supervision and real-time rendering on mobile devices, which are difficult to achieve with other representations such as neural radiance field \cite{weng2022humannerf, hu2023sherf, zhao2022humannerf} or 3D primitives \cite{li2024animatable, zheng2024gps, kocabas2024hugs} as they require more time-demanding rendering pipelines. However, above methods mostly focus on reconstructing static humans for the given poses, while lacking dedicated canonicalization and skinning for reconstructed avatars, thus are not suitable for animation-based applications \cite{wang2022arah}.

\begin{figure*}[htp!]
{\includegraphics[width=\textwidth]{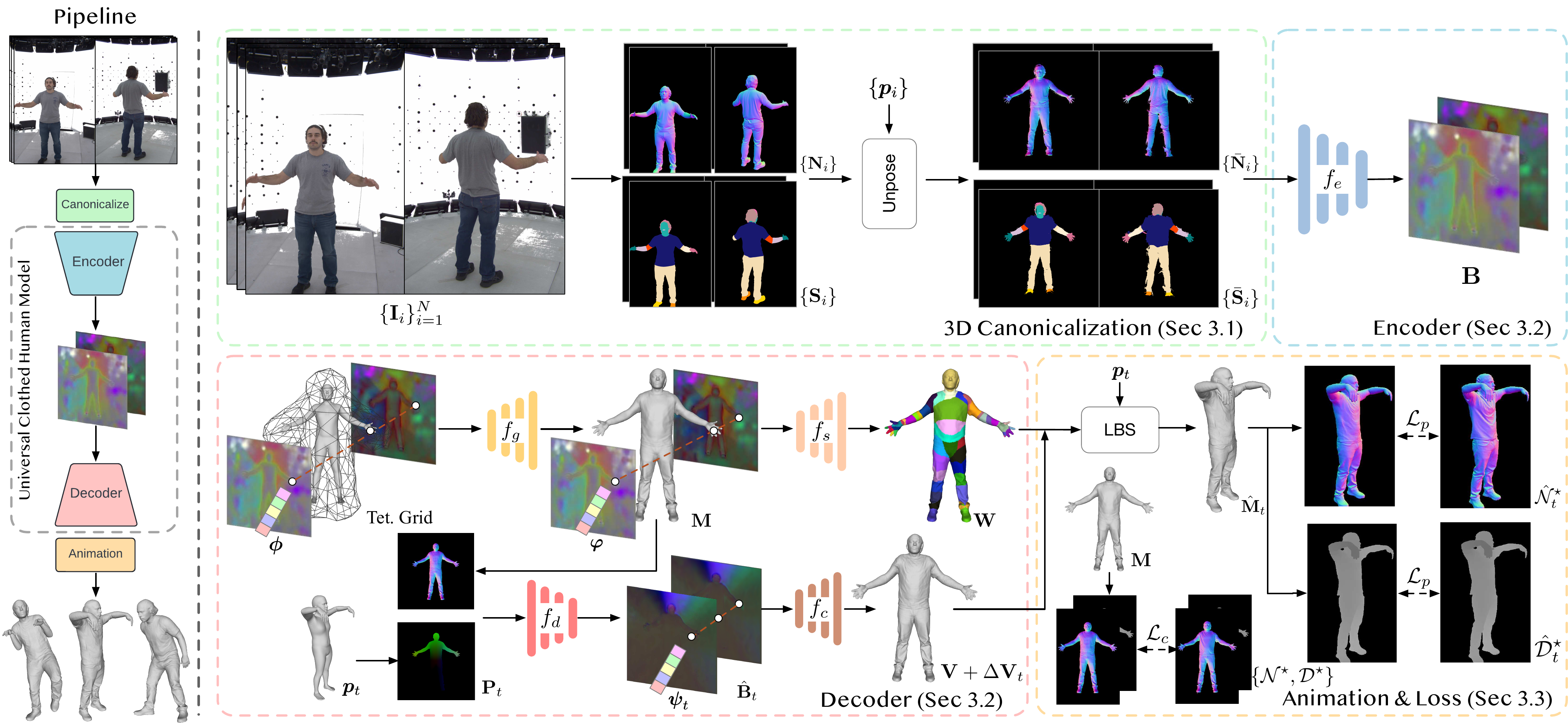}}
\vspace{-0.6cm}
    \centering\caption{\textbf{Method Overview. }We propose a novel method to feed-forwardly reconstruct personalized skinned avatars via a universal clothed human model. Specifically, given $N$ frames of posed human images $\{\mathbf{I}_i\}$ from front and back views, we first estimate their normal and segmentation images, and then unpose them for each frame and view to produce pixel-aligned initial conditions in a 3D canonicalization process (Section \ref{sec31}). Next, we propose to aggregate mult-frame references and produce a single bi-plane feature $\mathbf{B}$ as the representation of the subject identity. By sampling from this feature, we jointly decode personalized canonical avatar mesh $\mathbf{M}$, skinning weights $\mathbf{W}$ and pose-dependent vertex displacement ${\Delta} \mathbf{V}$  
    (Section \ref{sec32}) from a canonical tetrahedral grid. Finally, we adopt a multi-stage training process to train the model with posed-space ground truth and canonical-space regularization (Section \ref{sec33}). }
    \vspace{-0.1cm}
    \label{fig:main}
\end{figure*}

\noindent
\textbf{Animatable Avatar Reconstruction. }ARCH \cite{wang2022arah} pioneers the work to produce avatars that are more suitable for animation purposes. This is achieved by directly reconstructing avatars in a canonical space associated with a rigged template. Since it utilizes hand-crafted spatial encoding for geometric features, it tends to produce low-fidelity results with less accuracy and photorealism. ARCH++ \cite{he2021arch++} addresses this issue by learning geometry features from a PointNet++ \cite{qi2017pointnet++} encoder, but only uses a posed body template and thus loses fine-grained details. \cite{liao2023high} further refines the pipeline by sampling features from estimated normal images followed by post-processing for geometry refinement. Unlike these feed-forward approaches, \cite{ yang2024have, xiu2024puzzleavatar} rely on pretrained diffusion models \cite{liu2023zero1to3, rombach2022high} to hallucinate canonical avatar shapes from unconstrained photo collections, but require time-consuming per-subject optimization. Despite the improvement in avatar geometry, all these methods simply rig the avatar using skinning weights from nearest vertices of the rigged template for animation, which are only reliable for limited body shapes and cloth types. As a consequence, they often generate undesired deformation artifacts in challenging poses and extreme body shapes.

Recently, several works \cite{wang2022arah, saito2021scanimate, xue2023nsf, chen2024meshavatar, wang2021metaavatar, li2022tava} propose to jointly optimize personalized canonical shapes and skinning weights with pose-dependent deformation to reduce artifacts and improve animation quality. Specifically, \cite{wang2022arah, chen2021snarf} optimize correspondence between the canonical and posed space via an iterative root-finding algorithm \cite{chen2021snarf}. \cite{saito2021scanimate} assumes 3D scans as inputs and introduces an optimizable skinning field supervised by a cycle consistency loss. However, due to the lack of a universal prior for different body shapes and cloth types, these methods are restricted to a per-subject fitting approach and require computationally expensive test-time optimization to produce an avatar. \cite{wang2025towards} attempts to estimate personalized skinning weights and appear deformation on stylized characters, but requires known canonical character topologies. \cite{shin2024canonicalfusion} proposes to directly estimate subject-specific skinning maps from input images via a dual-encoder-decoder model. However, its canonicalization requires hand-crafted correction and thus is not end-to-end differentiable. In contrast to all above methods, we present a novel method that can jointly infer personalized geometry, skinning weights, and pose-dependent deformation all in an end-to-end learning framework.

\section{Method}
\textbf{Problem Definition. }Given $N$ frames of clothed human images $\{\mathbf{I}_i\}_{i=1}^{N}$ and estimated 3D body poses $\{\bm{p}_i\}_{i=1}^{N}$, we aim to produce an animatable 3D human avatar with personalized canonical geometry, skinning weights, and pose-dependent deformation for animation. Specifically, to ensure coverage of sufficient areas for the subject, we assume two images $\mathbf{I}_i^{f}$ and $ \mathbf{I}_i^{b} \in \mathbb{R}^{H \times W \times 3}$ from front and back views\footnotemark for each frame $i$, where $H$ and $W$ denote the image height and width. In addition, we assume the pose vector $\bm{p}_i$ consists of joint angles $\bm{\theta}_i \in \mathbb{R}^{J \times 3}$ and person-specific bone scales $\bm{s} \in \mathbb{R}^{J}$, both compatible with a pre-defined rigged body template with $J$ joints. For the output, we define the avatar geometry representation as a mesh $\mathbf{M}$ with vertices $\mathbf{V} \in \mathbb{R}^{V \times 3}$ and faces $\mathbf{F} \in \mathbb{R}^{F \times 3}$ in the canonical space, \emph{i.e.} aligned with the template. The vertices are associated with the joints as defined by a skinning weight matrix $\mathbf{W} \in \mathbb{R}^{V \times J}$, which allows the avatar to be animated by the LBS method. When animating the avatar with a target pose $\bm{p}_t$, we further predict pose-dependent vertex displacement $\Delta \mathbf{V}_t \in \mathbb{R}^{V \times 3}$ to correct LBS artifacts and improve animation quality. The method overview is shown in Figure \ref{fig:main} and we will introduce each module in the following sections.

\footnotetext{to ensure feasibility in phone photo setting, we allow each view to be taken at different time and thus do not need to be multi-view consistent.}

\subsection{3D Canonicalization}
\label{sec31}
Since posed human images can vary largely in body poses, scales and positions relative to the camera, directly extracting pixel features from raw images can be challenging and thus often leads to overly smooth geometry. To tackle this issue, we propose to normalize the variations in an explicit 3D canonicalization process with the following steps.

\noindent
\textbf{3D Lifting.} We first leverage an off-the-shelf human foundation model \cite{khirodkar2025sapiens} to independently estimate normal images $\mathbf{N}_i^v \in \mathbb{R}^{H \times W \times 3}$ for each frame and view $v$. Next, we follow \cite{xiu2023econ} to lift all normal images and produce front and back surface meshes via normal integration \cite{cao2022bilateral}. In addition, we also use \cite{khirodkar2025sapiens} to separately predict segmentation images $\mathbf{S}_i^v \in \mathbb{R}^{H \times W \times S}$, where $S$ denotes the number of class labels for each pixel. The segmentation images provide semantic guidance and help to produce locally consistent skinning weights. Finally, we back-project pixel segmentation labels onto lifted surface meshes as registered vertex attributes. We refer readers to more details of the lifting process in the supplementary materials. 

\noindent
\textbf{Mesh Unposing. }Once we have extracted posed surface meshes from input images, we align them in the shared canonical space via unposing, which is defined in Eq.(\ref{eq3}) as the inverse of the LBS operation as:
\begin{align}
    [\hat{\bm{u}}; 1] &= \text{LBS}(\bm{u}, \bm{w}, \mathbf{T}) = ({\textstyle\sum}_{j=1}^{J}\bm{w}_j\mathbf{T}_j) [\bm{u}; 1] \label{eq2} \;,\\ 
    [{\bm{v}}; 1] &= \text{LBS}^{-1}(\hat{\bm{u}}, \bm{w}, \mathbf{T}) = ({\textstyle\sum}_{j=1}^{J}\bm{w}_j\mathbf{T}_j)^{-1} [\hat{\bm{u}}; 1] \;, \label{eq3}
\end{align}

where $\hat{\bm{u}}, {\bm{u}} \in \mathbb{R}^3$ are a pair of posed and unposed vertices of a surface mesh, $[\cdot; 1]$ denotes a homogeneous vector, $\mathbf{T} \in \mathbb{R}^{J \times 4 \times 4}$ is the joint transformation computed from the pose vector (including the bone scales), and $\bm{w} \in \mathbb{R}^J$ is the associated vertex skinning weights. However, a key challenge for unposing is that the optimal skinning weight is unknown, as the personalized skinning weight is not available at this stage. While it is possible to first optimize it via a cyclic consistency loss \cite{saito2021scanimate}, such optimization can take hours to converge and thus is not desirable. Instead, we propose to use an arbitrary skinning weight, \emph{i.e.} from the nearest posed template vertices, and perform a deterministic unposing. The intuition is that while such unposing can produce noisy results, its output is only used an initial condition and the artifact patterns are consistent and thus can be later corrected by the universal model. We compare the effects of unposing in Figure \ref{fig:ab2} in the ablation study. 

\noindent
\textbf{Canonical Rendering. }For each unposed surface mesh, we use a fixed orthogonal camera $\bm{c}^v$ to render its unposed normal and segmentation images as $\bar{\mathbf{N}}_i^v$ and $\bar{\mathbf{S}}_i^v \in \mathbb{R}^{\bar{H} \times \bar{W} \times 3}$, where $\bar{H}$ and $\bar{W}$ are the height and width of the unposed images. Compared to the original posed images, body parts in unposed images are \emph{pixel-aligned} within each view (illustrated in Figure \ref{fig:main}), since we also normalize the bone scales in Eq.(\ref{eq3}). In this way, it significantly reduces the difficulty of feature extraction as features for each body vertex can be consistently queried from corresponding pixel locations, thus helps to generate canonical geometry with fine-grained details. In addition, projecting 3D surfaces onto 2D image space allows us to leverage powerful image models as shown in \cite{li2024animatable}, which further facilitates the feature extraction process. With the unposed normal and segmentation images as inputs, we then develop a universal clothed human model for reconstructing personalized skinned avatars as will be introduced in the following section.

\subsection{Universal Clothed Human Model}
\label{sec32}
To tackle the complexity of diverse body shapes and cloth types, we develop a universal model to learn a prior from a large collection of clothed human subjects. The model consists of an encoder that aggregates multi-frame initial conditions to refine canonicalization results and fuse a single feature representing the subject identity, followed by a decoder that jointly decodes personalized canonical shapes, skinning weights, and pose-dependent deformations.

\noindent
\textbf{Multi-Frame Encoder. }For challenging poses and cloth types, canonicalization artifacts can be challenging to correct due to the large deviation from the actual canonical shape. To tackle this issue, we propose to aggregate multiple posed references to improve model robustness and fuse a more plausible canonical avatar in a multi-frame encoder $f_e(\cdot)$. Specifically, we first stack unposed normal and segmentation images to align geometry and semantic references, and then follow \cite{trevithick2023real} to separately extract a high-resolution feature $\mathbf{H}_i^v$ for fine-grained geometric details, and a low-resolution feature $\mathbf{L}_i^v$ for global identity representation of the subject as:
\begin{equation}
    \mathbf{H}_i^v = f_h(\bar{\mathbf{N}}_i^v \oplus \bar{\mathbf{S}}_i^v), \quad \mathbf{L}_i^v = f_l(\bar{\mathbf{N}}_i^v \oplus \bar{\mathbf{S}}_i^v) \;,
\end{equation}

where $f_h(\cdot)$ is a shallow convolutional neural network (CNN) and $f_l(\cdot)$ is a DeepLabV3 \cite{chen2017rethinking} backbone, $\oplus$ denotes channel-wise concatenation. Unlike \cite{trevithick2023real}, we adopt more light-weight feature encoders and do not use vision transformers \cite{xie2021segformer}, thanks to the fact that the initial conditions are pixel-aligned. Finally, we aggregate features of all frames into a single bi-plane feature $\mathbf{B} \in \mathbb{R}^{\bar{H} \times \bar{W} \times 2C}$, with $C$ denoting the number of feature channels for each plane as:
\begin{equation}
    \mathbf{B} = (\mathbf{B}^f \oplus \mathbf{B}^b), \quad \mathbf{B}^v = \frac{1}{N}{\textstyle\sum}_{i=1}^N f_b(\mathbf{H}_i^v \oplus \mathbf{L}_i^v) \;,
\end{equation}

where $f_b(\cdot)$ is a CNN encoder. The feature aggregation is achieved by averaging features of each frame, which has several advantages. First, it flexibly supports an arbitrary number of input frames and is agnostic to the frame order. In addition, the averaged features favor details that are intrinsic to the subject, \emph{i.e.} body shapes and cloth types, while effectively disentangling pose-specific wrinkles and artifacts that do not commonly appear. We show in Figure \ref{fig:ab2} in the ablation study that the aggregated bi-plane feature allows the decoder to generate more plausible canonical geometry.

\noindent
\textbf{Canonical Geometry Decoder. }
To generate avatar meshes with arbitrary topology and fine-grained geometric details,  
we adopt a hybrid 3D representation DMTet \cite{shen2021deep}. Specifically, we first follow \cite{huang2024tech} to construct a tetrahedral grid encapsulated by an outer shell of the body template, as illustrated in Figure \ref{fig:main}. Such grid initialization allocates grid vertices mostly near the body template, thus helps to improve the geometry resolution. For each grid vertex $\bm{g} \in \mathbb{R}^3$, we orthogonally project it onto each plane in the bi-plane feature using the camera $\bm{c}^v$, and accordingly sample pixel features as $\bm{\phi}^v \in \mathbb{R}^{C}$. While $\bm{\phi}^v$ encodes the geometric and semantic information, we further incorporate the 3D position of the grid vertex as an inductive bias to encourage surface continuity and reduce floaters. Finally, we forward all features to a MLP geometry decoder $f_g(\cdot)$ to predict the SDF value $\bm{s} \in \mathbb{R}$ and vertex displacement $\Delta \bm{g} \in \mathbb{R}^3$ for each grid vertex as:
\begin{equation}
    (s, \Delta \bm{g})= f_g(\bm{\phi}^f \oplus \bm{\phi}^b \oplus \bm{g})\;.
\end{equation}

The canonical mesh $\mathbf{M}$ can then be differentiably extracted by Marching Tetrahedra (MT) \cite{shen2021deep} with the outputs.

\noindent
\textbf{Skinning Weight Decoder. }To ensure smooth animation for challenging poses and extreme body shapes, we propose to decode \emph{personalized} skinning weights in a separate MLP decoder $f_s(\cdot)$, in contrast to previous methods that rig the avatar with fixed skinning weights. Thanks to the decoded canonical avatar mesh, we can explicitly assign the skinning weight $\bm{w}$ for each canonical vertex $\bm{v}$. Similarly to the geometry decoder, we sample each canonical vertex from the bi-plane feature to obtain the skinning feature $\bm{\psi}^v \in \mathbb{R}^C$. To encourage neighboring vertices to have similar skinning weights, we also include the canonical vertex position $\bm{v}$ as the input feature as:
\begin{equation}
    \bm{w} = f_s(\bm{\varphi}^f \oplus \bm{\varphi}^b \oplus \bm{v}) \;,
\end{equation}

where the last layer in $f_s(\cdot)$ is a Softmax \cite{bridle1990probabilistic} layer to ensure the validity of the output skinning weights. 

\noindent
\textbf{Pose-dependent Deformation. }Since animation driven solely by LBS often leads to deformation artifacts, we propose to improve the animation quality by further including a pose-dependent deformation module as shape correction. Specifically, we first follow \cite{li2024animatable} to render front and back position map $\mathbf{P}_t^v \in \mathbb{R}^{\bar{H} \times \bar{W} \times 3}$ from the target pose vector $\bm{p}_t$ as the pose condition. We then combine it with the avatar identity condition, \emph{i.e.} the rendered decoded canonical mesh to obtain a deformation bi-plane feature $\hat{\mathbf{B}}_t \in \mathbb{R}^{\bar{H} \times \bar{W} \times 2C}$ as:
\begin{equation}
        \hat{\mathbf{B}}_t = (\hat{\mathbf{B}}^f_t \oplus \hat{\mathbf{B}}^b_t), \quad \hat{\mathbf{B}}^v_t = f_d(\mathbf{P}_t^v \oplus \mathcal{R}(\mathbf{M}, \bm{c}^v)) \;,
\end{equation}

where $f_d(\cdot)$ has the same architecture as the encoder $f_e(\cdot)$, and $\mathcal{R}(\cdot)$ is a renderer function to render normal images of canonical mesh with camera $\bm{c}^v$. Finally, we sample features $\bm{\psi}^v_t\in \mathbb{R}^{C}$ in a similar way as above and decode the per-vertex displacement $\Delta \bm{v} \in \mathbb{R}^3$:
\begin{equation}
    \Delta \bm{v}_t = f_c(\bm{\psi}^f_t \oplus \bm{\psi}^b_t) \;,
\end{equation}

where $f_c(\cdot)$ is a MLP decoder. With all above components, we can animate the avatar with any target joint transformation $\hat{\mathbf{T}}$ to produce the posed mesh $\hat{\mathbf{M}}_t$ as:
\begin{equation}
    [\hat{\bm{v}}_t;1] = \text{LBS}(\bm{v} + \Delta \bm{v}_t, \bm{w}, \hat{\mathbf{T}}) \;,\label{eq8}
\end{equation}

where $\hat{\bm{v}}_t$ is a vertex of the posed vertices $\hat{\mathbf{V}}_t \in \mathbb{R}^{V \times 3}$.

\subsection{Training Process}
\label{sec33} 
\noindent
Due to the coupling between the canonical geometry and skinning weights, supervising only with the ground truth in the posed space can generate avatars with incorrect components but can be accidentally warped to the correct target \cite{lin2022learning}. To address this issue, we propose a multi-stage training process to co-supervise the model with both posed-space ground truth and canonical-space regularization. 

\noindent
\textbf{Canonical-Space Stage.} We first train with only the biplane encoder $f_e(\cdot)$ geometry decoder $f_g(\cdot)$ as an initialization step. Since ground truth canonical meshes do not exist, we construct pseudo ground truths by unposing 3D scans of each frame in an optimization process (details are in the supplementary materials). Note the unposing here produces less artifacts but takes 20 minutes for each frame, thus is only applicable for data preparation. Next, we use a differentiable renderer \cite{pidhorskyi2024rasterized} to render the canonical mesh in a random set of $P$ views to produce the predicted canonical normal and depth images $\mathcal{N} \in \mathbb{R}^{P \times H \times W \times 3}$ and $\mathcal{D} \in \mathbb{R}^{P \times H \times W \times 3}$ respectively, and impose an $l_1$ loss as:
\begin{equation}
    \mathcal{L}_c = || \mathcal{N} - \mathcal{N}^{\star}_i ||_1 + || \mathcal{D} - \mathcal{D}^{\star}_i ||_1\;, \label{eq9}
\end{equation}

where $\mathcal{N}^{\star}_i$ and $\mathcal{D}^{\star}_i$ are the rendered normal and depth images of the pseudo ground truth canonical mesh in the same set of views. During training, we randomly select a frame $i$ in the input frame set for supervision, encouraging the geometry decoder to preserve details commonly appear in all frames, while discard artifacts that are frame-varying.

\noindent
\textbf{Posed-Space Stage.} Once the geometry decoder is sufficiently initialized by Eq.(\ref{eq9}), we then jointly train all modules by supervising in the posed space as:
\begin{equation}
        \mathcal{L}_p = || \hat{\mathcal{N}}_t - \hat{\mathcal{N}}^{\star}_t ||_1 + || \hat{\mathcal{D}}_t - \hat{\mathcal{D}}^{\star}_t ||_1 + \mathcal{L}_a (\hat{\mathcal{N}}_t, \hat{\mathcal{N}}^{\star}_t) \;, 
\end{equation}

where $\hat{\mathcal{N}}_t$ and $\hat{\mathcal{D}}_t$ are rendered images of predicted posed mesh, $\hat{\mathcal{N}}^{\star}_t$ and $\hat{\mathcal{D}}^{\star}_t$ are corresponding rendered images from ground truth posed scans, $\mathcal{L}_a$ is a perceptual loss \cite{zhang2018perceptual} to recover sharper wrinkles in the predicted posed mesh. In addition, we regularize the skinning weights prediction as:
\begin{equation}
    \mathcal{L}_s = || \mathbf{W} - \bar{\mathbf{W}}||_1 \;,
\end{equation}

where $\bar{\mathbf{W}}$ is the nearest skinning weights queried from the body template. Finally, we impose an edge loss
to penalize deformation artifacts of over-stretched triangles as:
\begin{equation}
    \mathcal{L}_e = \frac{1}{|\mathcal{E}|}{\textstyle\sum}_{\{i, j\} \in \mathcal{E}} \max(0, (\hat{\bm{v}}_i -  \hat{\bm{v}}_j)^2 - t) \;,
\end{equation}

where $\hat{\bm{v}}_i, \hat{\bm{v}}_j$ is a pair of vertex in the predicted posed mesh connected by an edge defined in $\mathcal{E}$, $|\mathcal{E}|$ denotes the total number of edges in the mesh, and $t$ is a constant threshold. The overall training objectives for the posed space training stage is thus a weighted sum of individual loss as $\mathcal{L} = \lambda_p \mathcal{L}_p + \lambda_s \mathcal{L}_s + \lambda_e \mathcal{L}_e$.
\section{Experiments}
\subsection{Dataset \& Metrics}
To train the universal model, we construct a large-scale dome capture dataset with 1100 clothed human subjects, with each subject performing up to 100 different poses. We leverage a surface reconstruction method \cite{cao2024supernormal} to extract 3D posed scans for each frame as the posed-space ground truths. More details are included in the supplementary material. We reserve 35 unseen subjects in our dataset (Dome Data) for testing. To evaluate model generalizability, we further render synthetic images from 130 subjects in RenderPeople \cite{rp} and directly test our pretrained models on it. We follow \cite{he2021arch++} to evaluate the geometric quality of reconstruction and animation results using 3 metrics:
average Point-to-Surface distance (P2S) and
Chamfer distance (CD) in centimeters, as well as L2 re-projected normal error (Normal) between the predicted and ground truth posed scans.

\begin{figure*}[htp!]
{\includegraphics[width=0.95\textwidth]{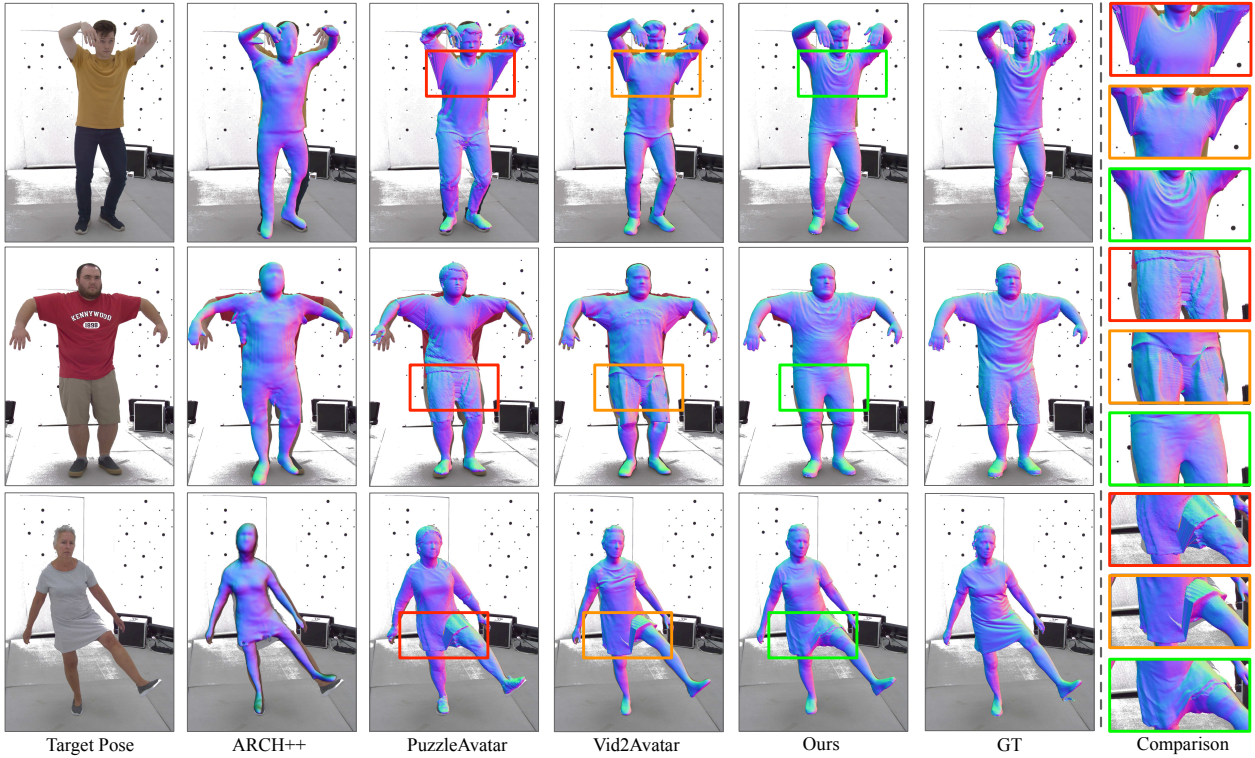}}
\vspace{-0.2cm}
    \centering\caption{\textbf{Qualitative Comparison. }Our method produces superior animation quality when reposed to an unseen pose for challenging poses, body shapes and cloth types, which reduces deformation artifacts, \emph{e.g.} stretched triangles, and generates plausible wrinkles.}
    \label{fig:quali}
    \vspace{-0.2cm}
\end{figure*}

\begin{figure*}[htp!]
{\includegraphics[width=0.95\textwidth]{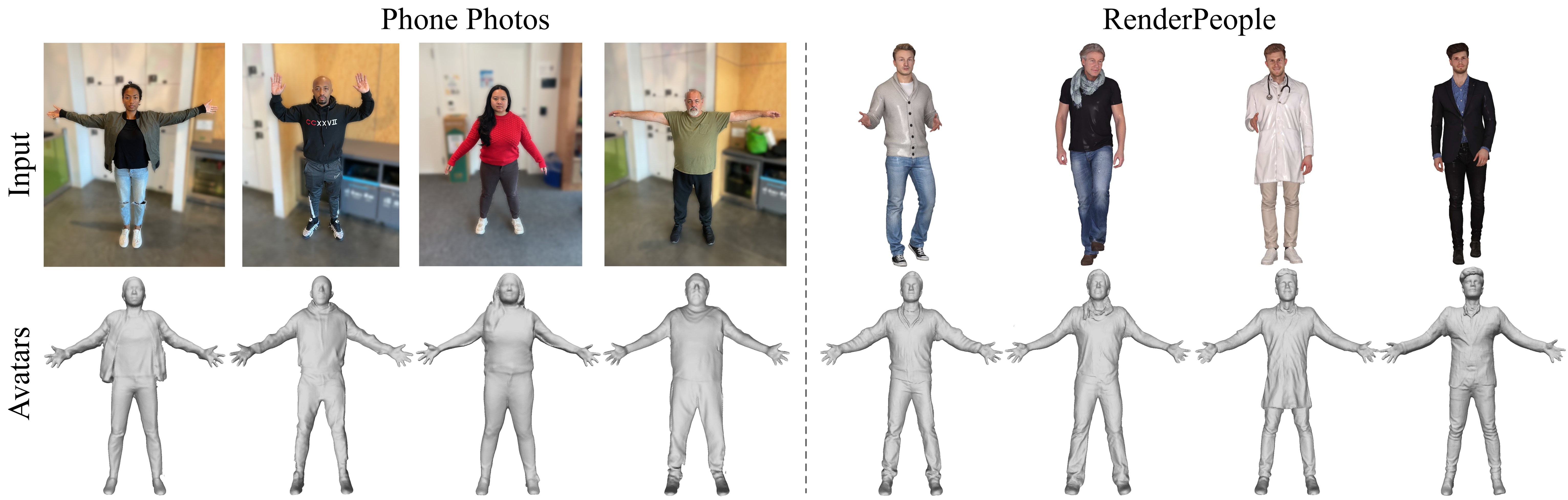}}
\vspace{-0.2cm}    \centering\caption{\textbf{Method Generalizability. }We show the pretrained universal model can directly apply to causally taken photos and synthetic images from Renderpeople \cite{rp}, which demonstrates its practical applications. When applied to phone photos, we do not require perfect alignment of front and back views and use \emph{estimated} poses from monocular images for canonicalization. More details are in appendix.}
    \label{fig:gene}
    \vspace{-0.5cm}
\end{figure*}

\begin{figure*}[htp!]
{\includegraphics[width=\textwidth]{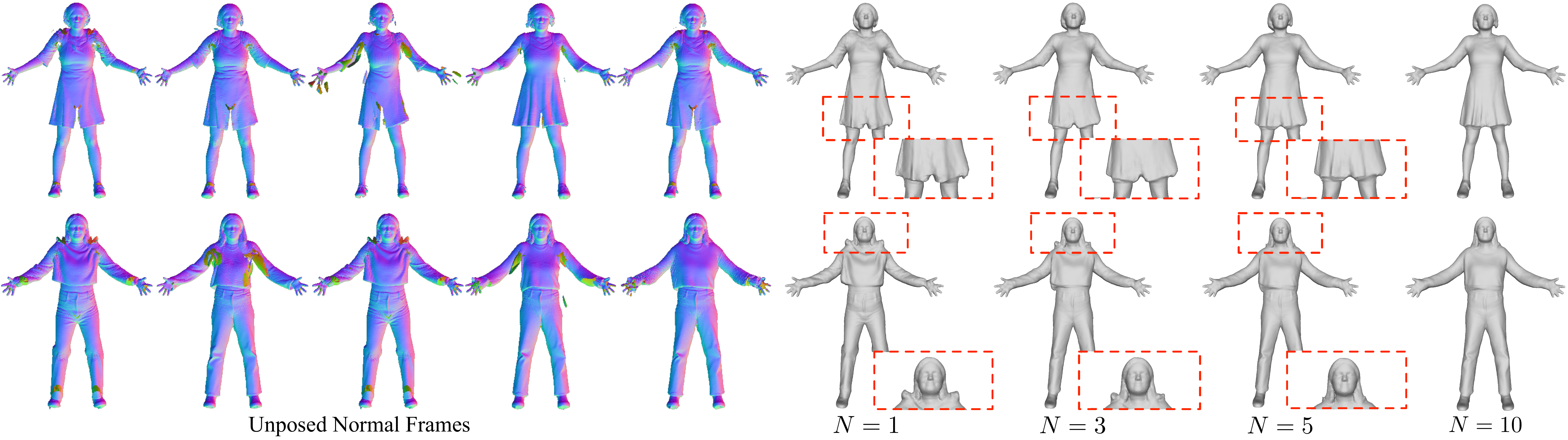}}
\vspace{-0.7cm}    \centering\caption{\textbf{Effects of multi-frame aggregation. }Given a set of unposed normal frames from different poses in the left, we show results of fused canonical shapes using the first $N$ frames at each column in the right. we observe that aggregation from multiple frames produces more plausible canonical shapes by correcting unposing artifacts, \emph{e.g.} on skirts and hairs, while preserving person-specific details. }
\vspace{-0.5cm}
    \label{fig:ab2}
\end{figure*}

\subsection{Implementation Details}
We implement our models in PyTorch \citep{paszke2017automatic} and perform all experiments on a single NVIDIA A100 GPU. We render all images with $ 512 \times 512$ pixels. We train the model using the Adam \citep{kingma2014adam} optimizer for 10K iterations in the canonical-space stage, and 100K iterations in the posed-space stage, both with a learning rate of $1 \times 10^{-4}$. The total training process takes 2 days to converge, and we report the inference time comparison in Table \ref{table1}. For the training losses, we set $\lambda_{p} = 1$, $\lambda_s = 0.1, \lambda_e = 100$ respectively, and use an edge length threshold of $t = 1 \times 10^{-4}$. More details and network architecture are included in the appendix.

\subsection{Results and Comparisons}
We compare with three types of baselines: (\emph{i}) learning-based method \cite{he2021arch++} that produces animatable avatars in a feed-forward approach, (\emph{ii}) optimization-based method \cite{guo2023vid2avatar} that optimizes canonical avatar shapes from multi-view videos, and (\emph{iii}) diffusion-based method \cite{xiu2024puzzleavatar} that generates avatar with SDS loss \cite{poole2022dreamfusion}. As all baselines do not predict personalized skinning weights, we follow their implementation to animate the avatar with a fixed skinning weight (queried from nearest template vertices). For \cite{guo2023vid2avatar, xiu2024puzzleavatar} and our method, we reconstruct the canonical avatar from the same set of posed images, and test the quality of animation given an unseen target pose. For a fair comparison, we adapt both baselines to use the same rigged template as us and provide them with ground truth pose vectors. Since \cite{he2021arch++} use only a single input image and do not have training code, we are unable to fairly compare with it and thus only test its reconstruction quality using the target frame with its own rigged template. In Figure \ref{fig:ab1}, we implement a similar sampling strategy with it and compare with our canonicalization approach instead. More comparison with other reconstruction and animation methods are included in the appendix.
\vspace{-0.6cm}
\begin{table}[htp!]
\centering
\caption{\textbf{Quantitative comparison with existing methods.} Our method achieves superior geometry quality than existing methods \citep{he2021arch++, xiu2024puzzleavatar, jiang2022selfrecon}, and requires significantly less inference time. }
\vspace{-0.2cm}
\label{tab:garment_deformation}
\resizebox{0.48\textwidth}{!}{
\begin{tabular}{@{}lccccccc@{}}
\toprule
\multirow{2}{*}{Methods} & \multicolumn{3}{c}{Dome Data} & \multicolumn{3}{c}{RenderPeople}  & \multirow{2}{*}{Time} \\ \cmidrule(l){2-7} 
                 & Normal $\downarrow$ & P2S $\downarrow$ & CD $\downarrow$ & Normal $\downarrow$ & P2S $\downarrow$ & CD $\downarrow$  \\ \midrule
ARCH++ \citep{he2021arch++} & 0.338 & 4.52 & 5.07 & 0.195 & 2.44 & 2.60 & 26s \\
Vid2Avatar \cite{guo2023vid2avatar} & 0.072 & 0.98 & 1.12 & - & - & - & 8h \\
PuzzleAvatar \citep{xiu2024puzzleavatar} & 0.104 & 1.47 & 1.63 & 0.132 & 1.79 & 1.91 & 3h\\ \hline
Ours (LBS Only) & 0.030 & 0.43 & 0.49 & 0.026 & 0.33 & 0.38 & 18s \\ 
Ours (Full Model) & \textbf{0.026} & \textbf{0.37} & \textbf{0.43} & \textbf{0.024} & \textbf{0.30} & \textbf{0.34} & \textbf{18s} \\
\bottomrule
\end{tabular}
}
\label{table1}
\vspace{-0.3cm}
\end{table}

As compared in Table \ref{table1}, our method outperforms baseline methods by a large margin in all geometry metrics, demonstrating results with superior geometric quality. Moreover, in contrast to \cite{guo2023vid2avatar, xiu2024puzzleavatar} that require hours of test-time optimization, we can produce personalized skinned avatars in less than 20 seconds (time details in the supplementary materials) thanks to the feed-forward reconstruction. We further compare qualitative results with baselines in Figure \ref{fig:quali}. We observe the fixed skinning strategy inevitably leads to deformation artifacts, \emph{e.g.} over-stretched triangles, for challenging poses and body shapes. In comparison, we generate more smooth and realistic animation and effectively reduces deformation artifacts for even loose garments like skirts. We also observe that diffusion-based method \cite{xiu2024puzzleavatar} can fail to preserve facial identity, while our method faithfully reconstructs personalized details. Moreover, we show that our pose-dependent deformation module produces plausible wrinkle patterns, while \cite{guo2023vid2avatar} simply bakes wrinkles in canonical shapes without further deformations. This module also helps to reconstruct more accurate geometric details as compared in Table \ref{table1}. Finally, we show in Figure \ref{fig:gene} that the universal prior directly applies to phone photos or synthetic characters.

\subsection{Ablation Study}

\begin{figure}[htp!]
{\includegraphics[width=0.45\textwidth]{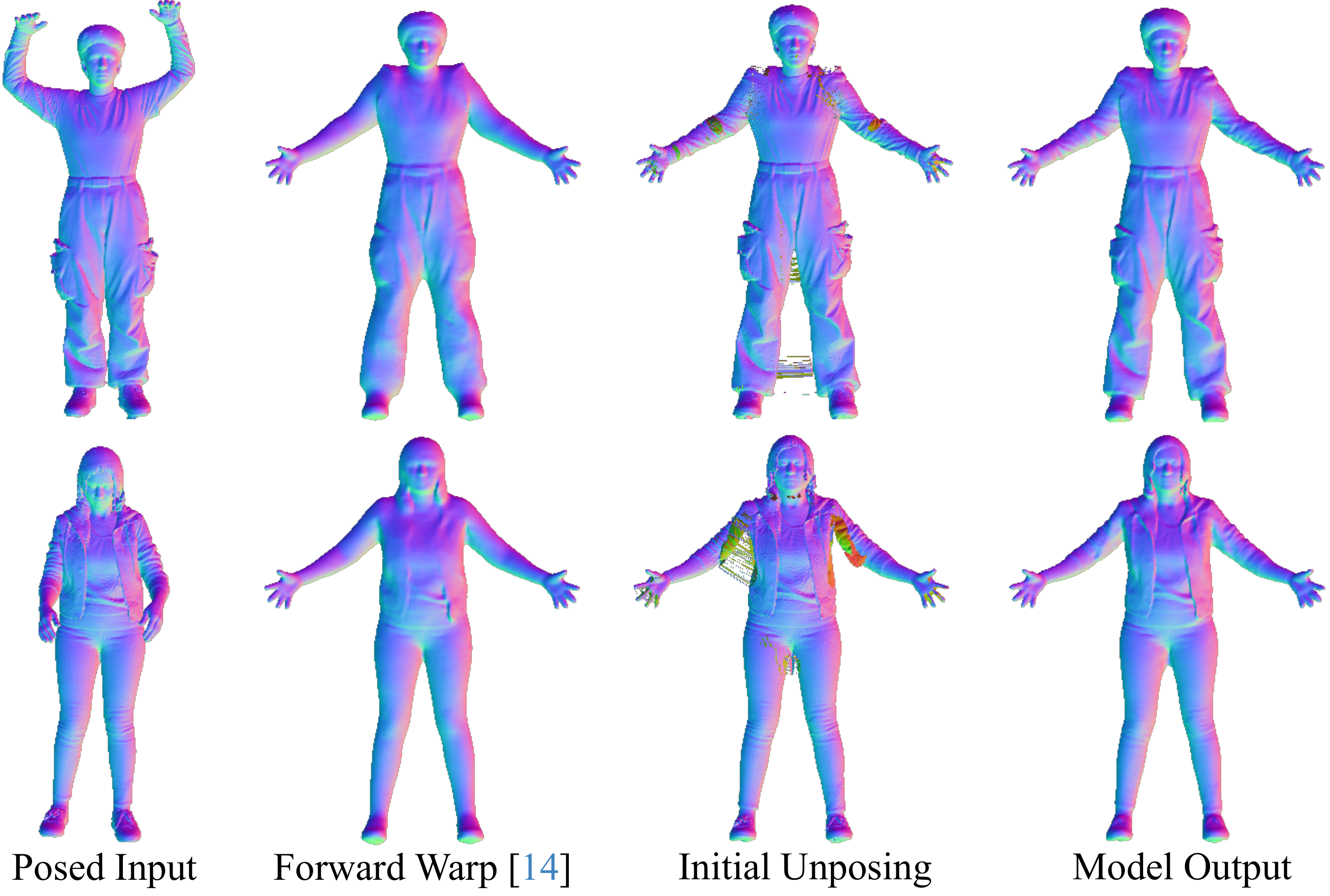}}
\vspace{-0.3cm}
    \centering\caption{\textbf{Effects of canonicalization. }By taking canonicalization results as inputs, the universal model can learn to reduce unposing artifacts and preserve fine-grained details, compared to directly sampling features from posed inputs via forward warping as \cite{he2021arch++}.}
    \label{fig:ab1}
    \vspace{-0.2cm}
\end{figure}

\noindent
\textbf{Effects of Canonicalization. }We show the effects of canonicalization by comparing the results of reconstructed canonical geometry using one posed reference in Figure \ref{fig:ab1}. Since input subjects vary largely in poses and shapes, directly sampling from posed inputs, \emph{i.e.} by warping query vertices via forward LBS similar to \cite{he2021arch++}, retrieves misaligned features and thus leads to overly smoothed geometry (second column). In contrast, the canonicalization can produce pixel-aligned initial conditions. However, due to the inaccurate skinning, the initial unposing results often contain artifacts, \emph{e.g.} stretched triangles, that are not suitable for use. To tackle this, we refine them in the universal model to effectively refine a more plausible canonical shape, while preserving fine-grained geometric details.

\begin{figure}[htp!]
\vspace{-0.2cm}
{\includegraphics[width=0.46\textwidth]{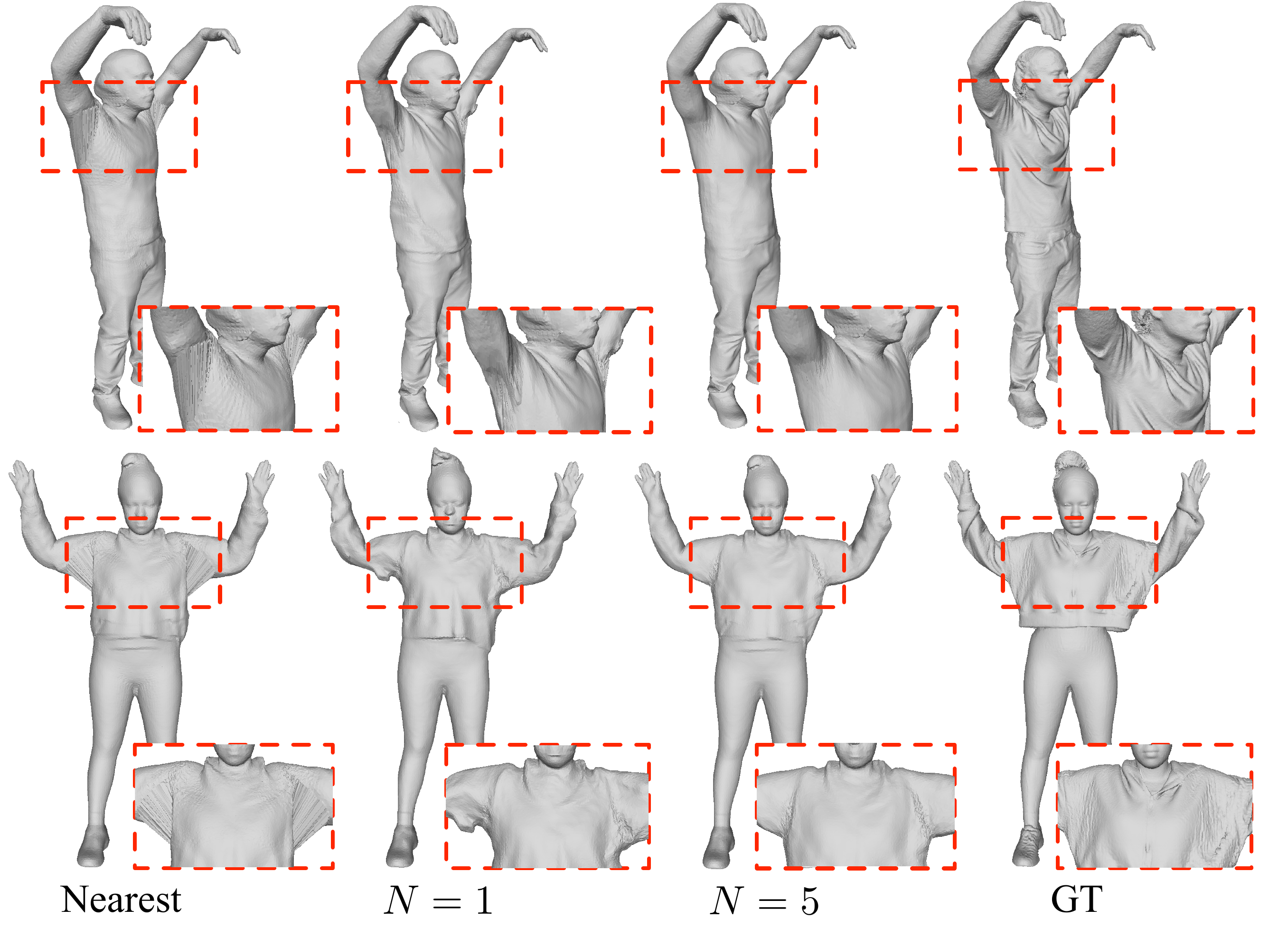}}
\vspace{-0.2cm}
\centering\caption{\textbf{Effects of personalized skinning weights. }We show personalized skinning weights reduce deformation artifacts, \emph{e.g.} under armpit, and can be more robustly estimated when trained with multiple input and target frames. Note we show results deformed by LBS only, \emph{i.e.} without pose-dependent deformation.}
    \label{fig:ab3}
    \vspace{-0.7cm}
\end{figure}

\noindent
\textbf{Effects of Multi-Frame Aggregation. }We compare the reconstructed canonical shapes with different numbers of reference frames in Figure \ref{fig:ab2}. Due to the lack of ground truth canonical shapes, refinement for canonicalization using a single frame can be challenging for loose garments and long hairs when input poses are not ideal. In comparison, by including more frames, the multi-frame encoder learns to preserve intrinsic features that commonly appear across different frames, thus can fuse a more plausible avatar. In practice, we observe the fusion results saturate in 5 frames to produce sufficiently plausible canonical shapes.

\noindent
\textbf{Effects of Personalized Skinning Weights. }We show the benefits of predicting personalized skinning weights in Figure \ref{fig:ab3}. In particular, skinning the avatar using nearest template body vertices can produce over-stretched triangles, \emph{e.g.} under armpit, while personalized skinning weights effectively reduces such deformation artifact. Furthermore, we show that training with multiple input and target frames ($N = 5$) produces more robust skinning weights than only learn to repose to the same frame as input ($N = 1$).

\begin{figure}[htp!]
{\includegraphics[width=0.45\textwidth]{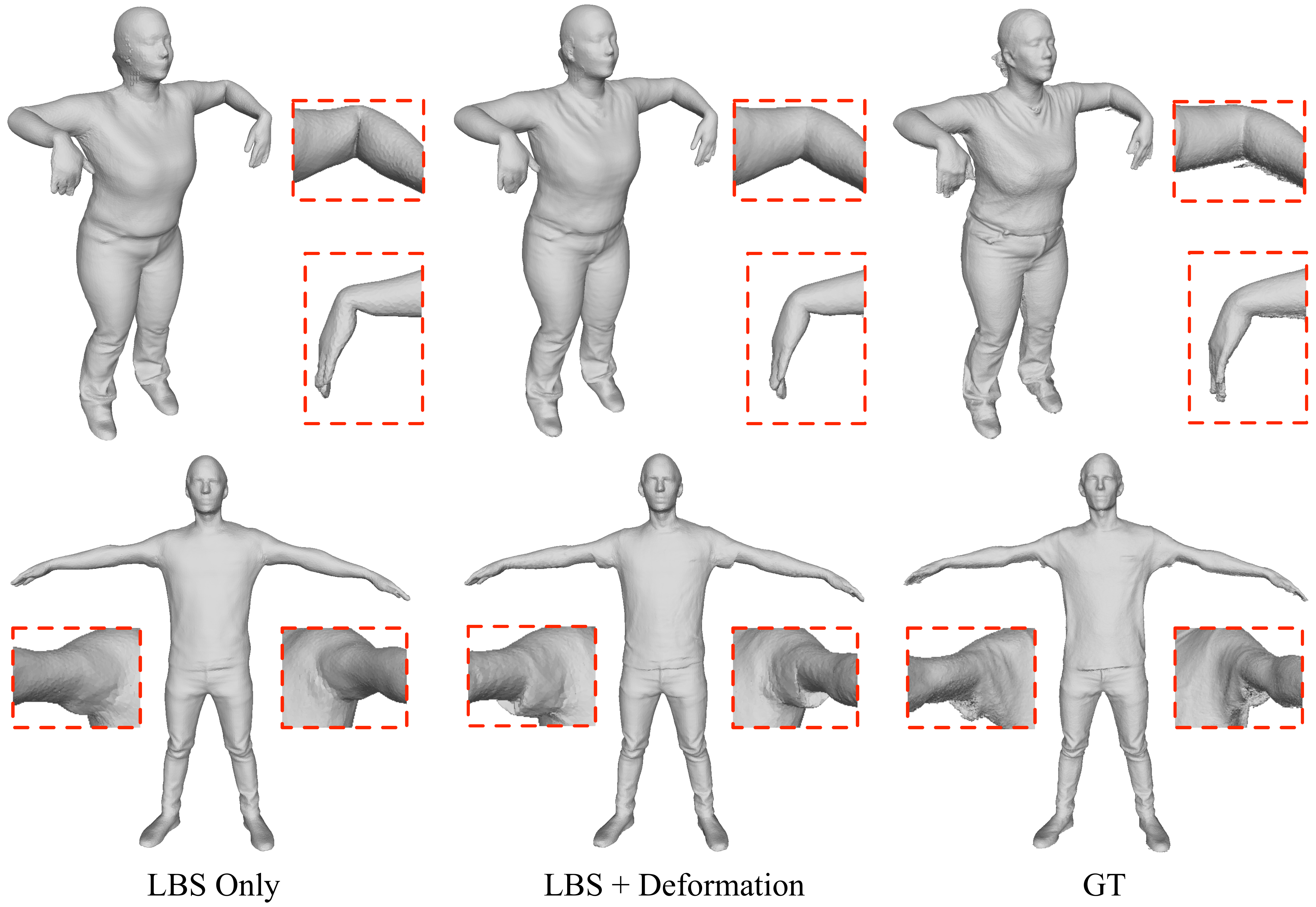}}
\vspace{-0.2cm}    \centering\caption{\textbf{Effects of Pose-dependent Deformation. }The deformation module can correct LBS artifacts (wrist and arm bending in first row) and generate plausible garment dynamics (sleeves draping in second row), which improves animation realism. }
    \label{fig:ab4}
    \vspace{-0.3cm}
\end{figure}

\noindent
\textbf{Effects of Pose-dependent Deformation. }As shown in Figure \ref{fig:ab4}, we observe three benefits on including pose-dependent deformation. First, it corrects LBS artifacts, \emph{e.g.} when bending elbow and wrist, which improves animation realism. Second, it can generate plausible garment dynamics such as sleeves draping when raising  arms. Finally, we observe an overall refinement of fidelity in wrinkles details.

\section{Discussion}
\textbf{Limitation.}
Although our method achieves superior reconstruction results, the geometric fidelity of the avatars are bounded by the resolution of the tetrahedral grids, thus may lead to missing details such as tiny accessories. Moreover, we only consider deformations dictated by pose, omitting complex dynamics such as body-cloth interactions, and motions of long hairs or extremely loose garments whose deformations are not entirely pose-dependent.

\noindent
\textbf{Conclusion.} In this paper, we present FRESA, a novel method for feed-forward reconstruction of personalized skinned avatars using only a few images. The core to our method is a universal clothed human model learned from diverse human subjects, which achieves joint inferring of personalized canonical shapes, skinning weights, and pose-dependent deformations. Moreover, we introduce a 3D canonicalization process with multi-frame aggregation to further improve the model robustness. Thanks to these innovations, our method achieves instant generation and zero-shot generalization with superior animation quality.
\clearpage
\setcounter{page}{1}

\section*{A. Data Acquisition}

\begin{figure}[htp!]
\vspace{-0.2cm}
{\includegraphics[width=0.48\textwidth]{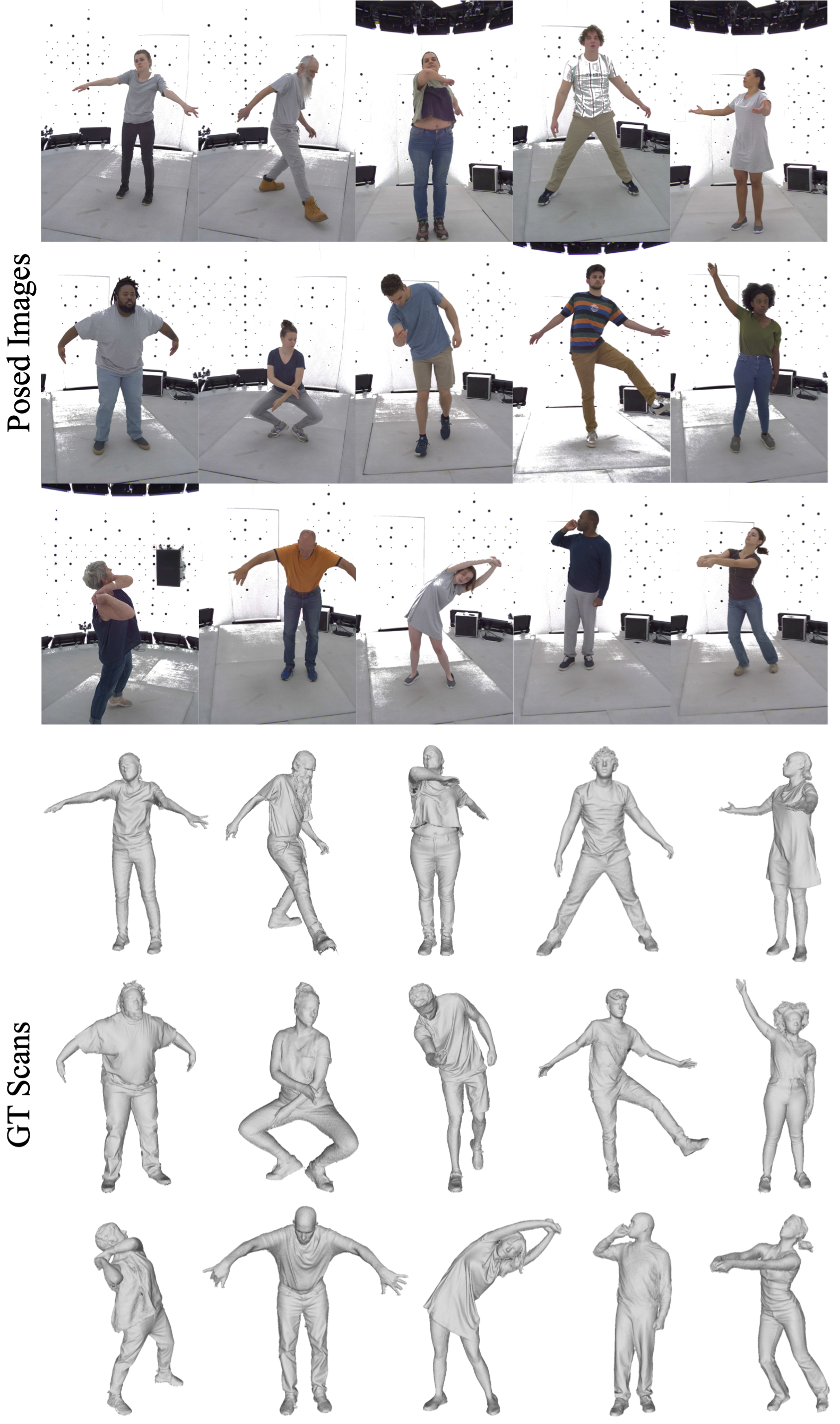}}
\vspace{-0.8cm}
    \centering\caption{\textbf{Samples of dome data. }Our dataset contains diverse posed clothed humans paired with high-quality 3D scans as ground truths, which facilitates learning an effective universal prior.}
    \label{fig:data}
\end{figure}

\noindent
\textbf{Data Capture. }We visualize our dome capture data in Figure \ref{fig:data}. For each subject, we capture 128 images
from different view points using a group of fixed cameras, and adopt SuperNormal~\cite{cao2024supernormal} to reconstruct the 3D scans for geometry supervision. We further estimate 3D joints for each frame from multi-view images, and adopt an incremental pose encoder \cite{carreira2016human} to obtain the pose vector $\bm{p}$. With the diverse posed clothed humans and high-quality scans, we can learn an effective universal prior that well generalizes to phone photos.

\begin{figure}[htp!]
\vspace{-0.2cm}
{\includegraphics[width=0.48\textwidth]{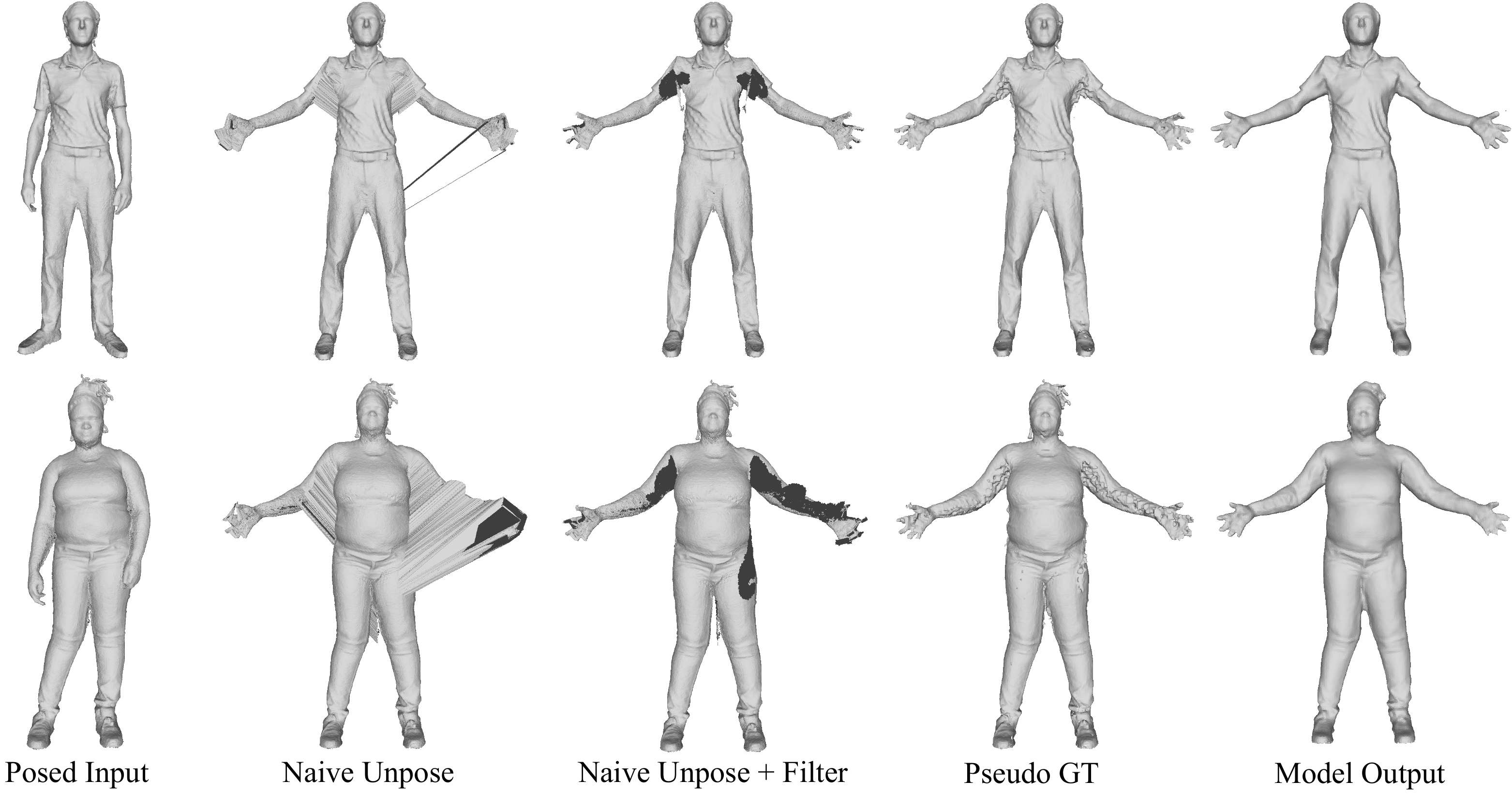}}
\vspace{-0.5cm}
    \centering\caption{\textbf{Unposing Comparison. }We compare the results between naive unposing (used in the inference pipeline) and pseudo GT via optimization (used for data preparation). The second approach produces more plausible results but requires significantly more time. Note we filter edges with length larger than $1 \times 10^{-4}$ to reduce noises. }
    \label{fig:flexi}
    \vspace{-0.2cm}
\end{figure}

\noindent
\textbf{Pseudo-GT Canonical Meshes. }To resolve the coupled ambiguity between canonical shapes and skinning weights, we construct pseudo-GT\footnote{ideal GT meshes should be obtained by unposing scans with personalized skinning weights, which is not available in canonical-space stage.} canonical meshes as the regularization for canonical-space stage training. Specifically, we adopt FlexiCubes \cite{shen2023flexible} as the 3D geometry representation and build a cubic grid with $G = 256^3$ vertices near the rigged body template. We then initialize the SDF value $s$ of each grid vertex as the signed distance to the template, and optimize canonical vertices $\bar{\mathbf{V}}$ parameterized by the Flexicube parameters $s, \bm{\alpha}, \bm{\beta}, \bm{\gamma}$ \cite{shen2023flexible} such that:
\begin{equation}
    \bar{\mathbf{V}} = \arg\min_{\mathbf{V}} (||\mathcal{R}_{d} (\Tilde{\mathbf{V}}) - \hat{\mathcal{D}}^{\star}||_1 + ||\mathcal{R}_{n} (\Tilde{\mathbf{V}}) - \hat{\mathcal{N}}^{\star}||_1 + \mathcal{L}_r) \;,
\end{equation}

where $\mathcal{R}_{d}(\cdot)$ and $\mathcal{R}_{n}(\cdot)$ are renderer functions for normal and depth rendering, $\Tilde{\mathbf{V}}$ is the posed vertices obtained by forward LBS (where skinning weights queried from nearest template vertices), $\mathcal{L}_r$ is the regularization term in \cite{shen2023flexible}, and $\hat{\mathcal{D}}^{\star}, \hat{\mathcal{N}}^{\star}$ are ground truth depth and normal images for posed scans. We empirically observe that such unposing strategy reduces artifacts of over-stretched triangles as shown in Figure \ref{fig:flexi}. However, since such unposing requires a complete optimization process and takes 20 minutes to converge, it is only suitable for data preparation. 

\noindent
\textbf{Approval of Usage. }All participants involved in dome capture data and phone photos have signed a consent form that authorize the usage of their images for model development and academic publications.

\section*{B. Canonicalization Details}
\label{secb}
\textbf{Pose Tracking for Photos. }We use an artists designed rigged body mesh as the template, which contains $J = 67$ joints. For each front and back view photo, we estimate its 2D joint positions using \cite{khirodkar2025sapiens}, and optimize the pose vector $\bm{p}$ to minimize the 2D projection loss similar to \cite{bogo2016keep}. We further determine the absolute scale of the subject based on a pretrained statistical prior model using PCA coefficients. The overall optimization process takes about 1 minute per frame. For a fair comparison, in the main paper, we report inference time for all methods excluding the pose tracking time as we assume known poses in our pipeline.

\begin{figure}[htp!]
{\includegraphics[width=0.46\textwidth]{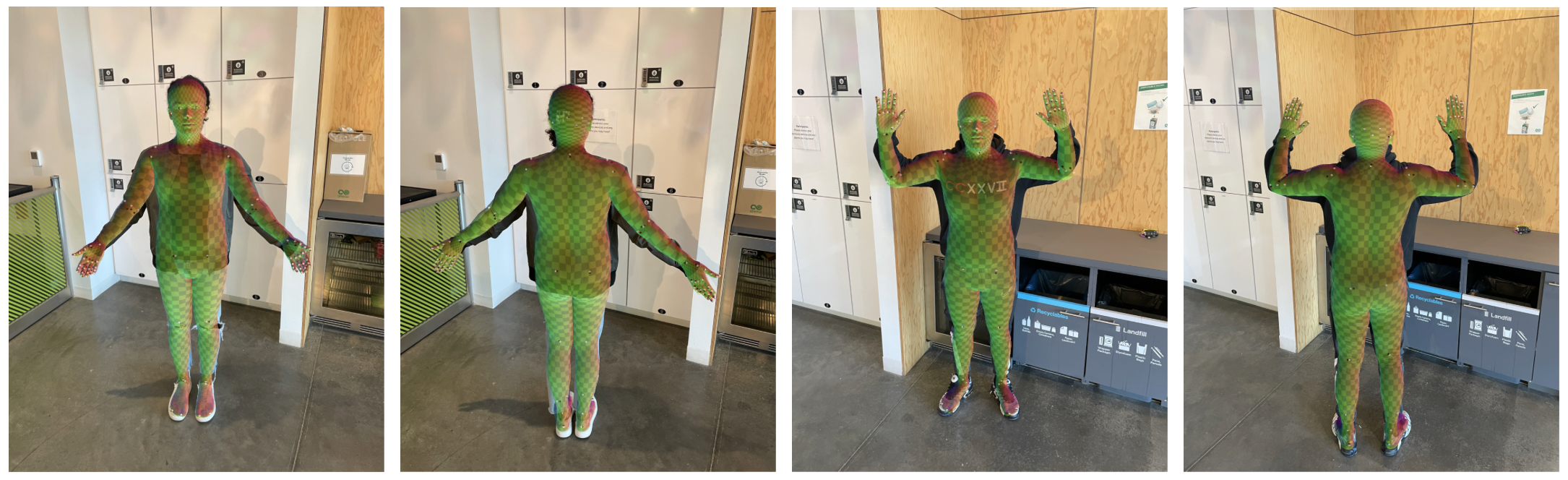}}
\vspace{-0.1cm}
\centering\caption{\textbf{Illustration of settings for photos. } We use \emph{estimated} body poses and do not require perfect alignment between views. }
    \label{fig:rebuttal1}
\end{figure}

Note that to ensure a practical use of our method, we do not require a perfect alignment between front and back views for causally taken photos, \emph{i.e.} we do not require known camera poses, and photos do not need to be synchronized in time, as illustrated in Figure \ref{fig:rebuttal1}. Such casual inputs can be robustly handled with the universal clothed human prior and multi-frame aggregation, as shown in main paper. In addition, since there are no GT body poses for phone photos, we use an off-the-shelf pose estimator to estimate body poses for each view. Demo results in the main paper show that our method can robustly generate plausible avatars under this imperfect unposing, ensuring a practical use of our method. Moreover, the proposed multi-frame aggregation approach can further improve robustness against inaccurate pose estimation in individual frame.

\begin{figure}[htp!]
{\includegraphics[width=0.48\textwidth]{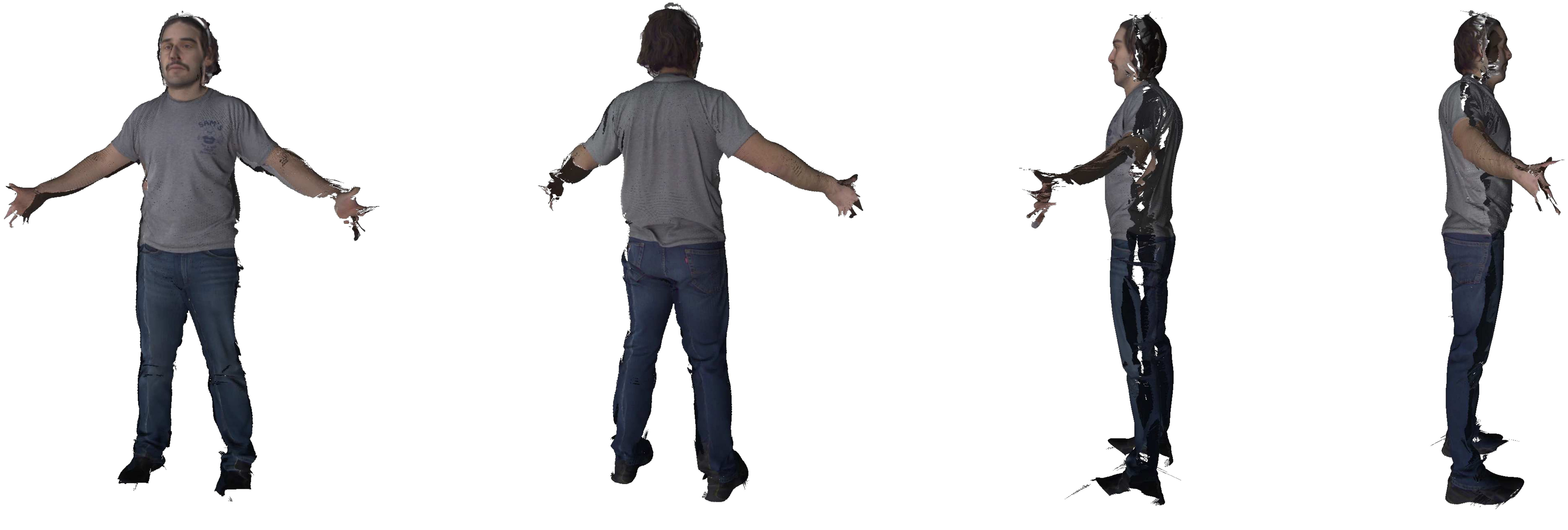}}
\vspace{-0.8cm}
    \centering\caption{\textbf{Illustration of Lifted Surface Meshes. }Note we removed the over-stretched edges after unposing. The lifting process produces two unposed surface meshes but can not be perfectly aligned in boundary. }
    \label{fig:ab_dbini1}
    \vspace{-0.2cm}
\end{figure}

\noindent
\textbf{3D Lifting. }We follow \cite{xiu2023econ} to use d-BiNI method to obtain the lifted front and back surface meshes for each frame. The surface depth is initialized based on the tracked poses, \emph{i.e.} the surface depth of posed body template. We visualize the resulting unposed surface meshes in Figure \ref{fig:ab_dbini1}.

\begin{figure}[htp!]
\vspace{-0.2cm}
{\includegraphics[width=0.48\textwidth]{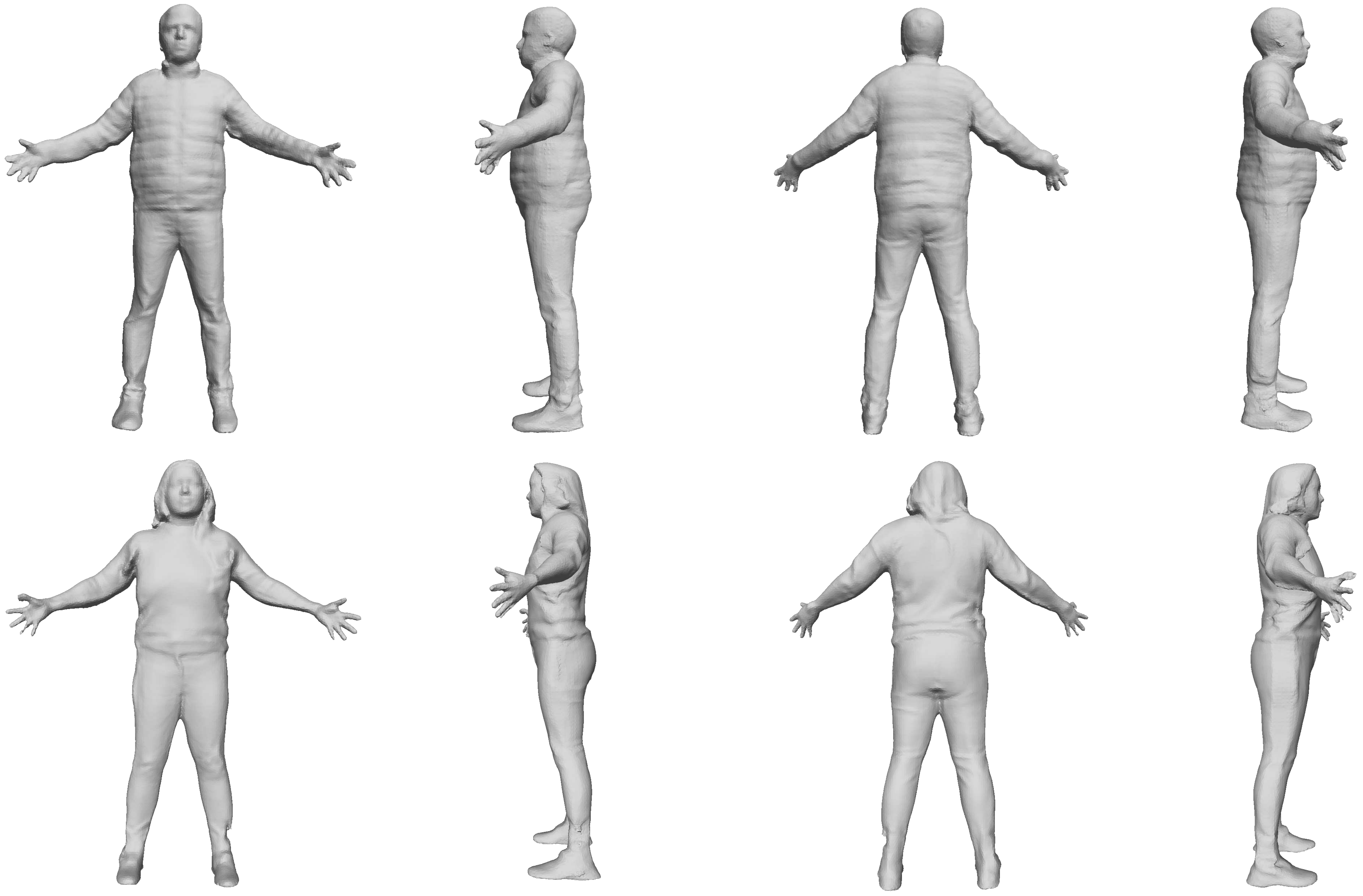}}
\vspace{-0.5cm}
    \centering\caption{\textbf{Visualization in Four Views. }By only taking inputs of front and back views, our method can infer plausible side-view geometry and produce a consistent boundary. }
    \label{fig:four}
    \vspace{-0.2cm}
\end{figure}

\begin{figure}[htp!]
\vspace{-0.2cm}
{\includegraphics[width=0.48\textwidth]{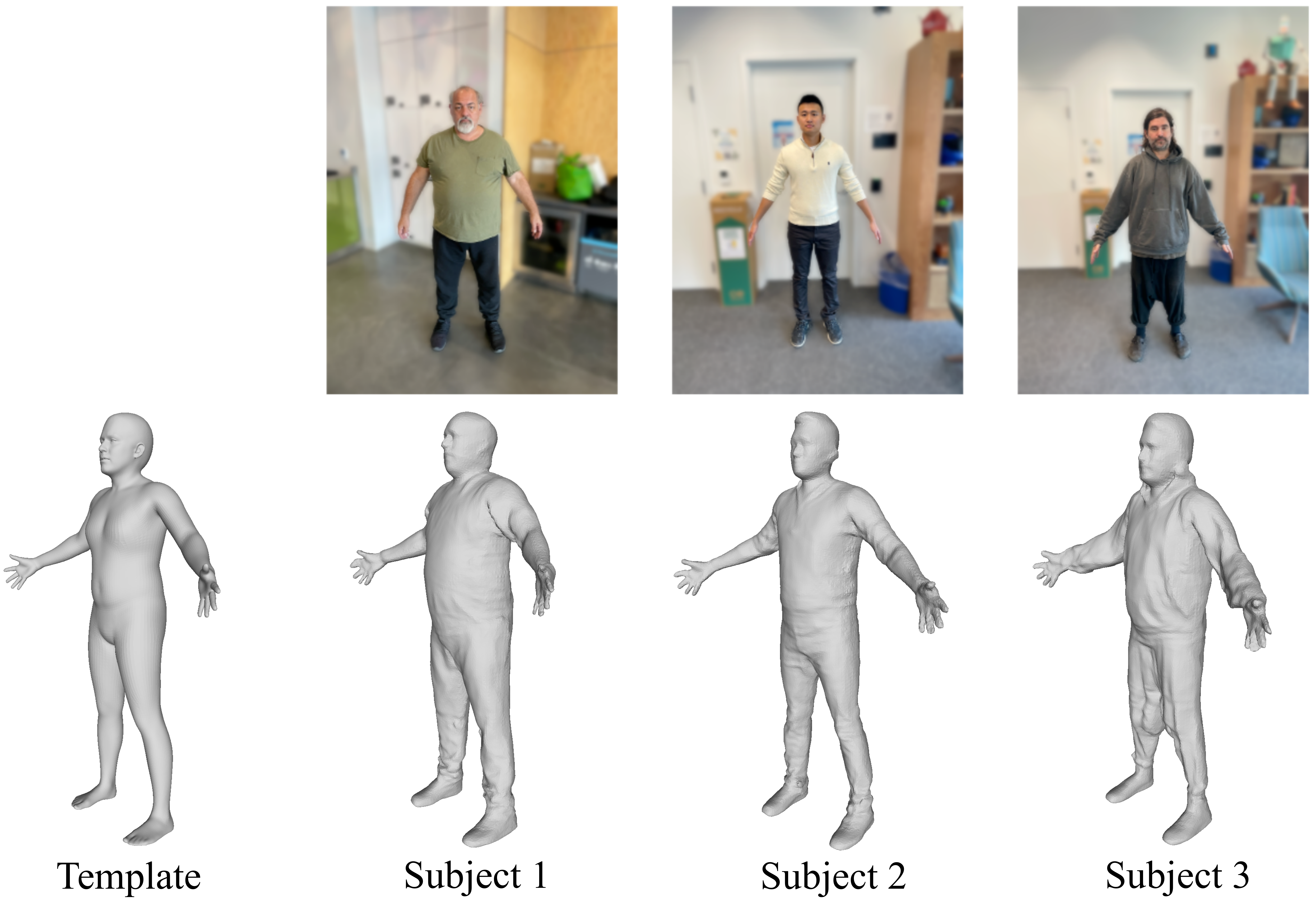}}
\vspace{-0.5cm}
    \centering\caption{\textbf{Results of Inferred Body Shape. }Our method can produce personalized body shapes based on input conditions and is not restricted to the template shape. }
    \label{fig:ab_dbini2}
    \vspace{-0.2cm}
\end{figure}

In contrast to \cite{xiu2023econ} that attempts to directly complete the lifted meshes in the 3D space, we re-render them into 2D images as initial conditions, and infer the canonical shape \emph{from scratch} for two reasons: (\emph{i}) it produces a more plausible boundaries by jointly refining geometry in both visible and invisible parts, as shown in Figure \ref{fig:four}. (\emph{ii}) it  can learn a personalized body shape instead of a fixed shape bounded by the initial depth, as shown in Figure \ref{fig:ab_dbini2}. Finally, we choose to use two views in the paper as the lifting and pose tracking process work mostly robust in these two views. However, our method can still produce a plausible side view geometry by learning across diverse subjects.

\section*{C. More Implementation Details}

We train our model using $N = 5$ input frames, which achieves the best balance between plausibility and fidelity. In inference, our model can be applied to an arbitrary number of input frames based on availability. To increase the model generalizability, we apply a data augmentation by mixing unposing results from both d-BiNI and 3D scans when training the model. Finally, we use \cite{quartet} to tetrahedralize the volume near the canonical template (with a distance of 0.2m), resulting in a grid of resolution 256. More details about our network architecture are included in Appendix D.

\noindent
\textbf{Inference Time Details. }We report the inference time for one input photo with image size $1280 \times 960$. Specifically, (\emph{i}) the segmentation and normal estimation for \cite{khirodkar2025sapiens} takes 4.90s, (\emph{ii}) the d-BiNI time for both views takes 9.91s, (\emph{iii}) the unposing (including finding the nearest template vertices) takes 1.54s, (\emph{iv}) the canonical rendering takes 0.06s, and (\emph{v}) the overall model inference takes 1.64s, thus the total inference time is 18.05s. All time are reported with a single NVIDIA A100 GPU. 

\noindent
\textbf{Baseline Implementation. }For \cite{he2021arch++}, we use the test code and pretrained models provided by the author. Since the method only takes a single image as input, we test its reconstruction quality by using the target posed image as input. For \cite{guo2023vid2avatar, xiu2024puzzleavatar}, we modify its code to use our rigged template and canonical pose instead of SMPL-X \cite{pavlakos2019expressive} template. For \cite{guo2023vid2avatar}, we also follow their implementation to use nearest template vertices’ skinning weights,
weighted by the point-to-point distances in deformed space, while \cite{xiu2024puzzleavatar} only uses a template mesh to initialize the DMTet grid.

\section*{D. Network Architecture}
\begin{figure*}
{\includegraphics[width=\textwidth]{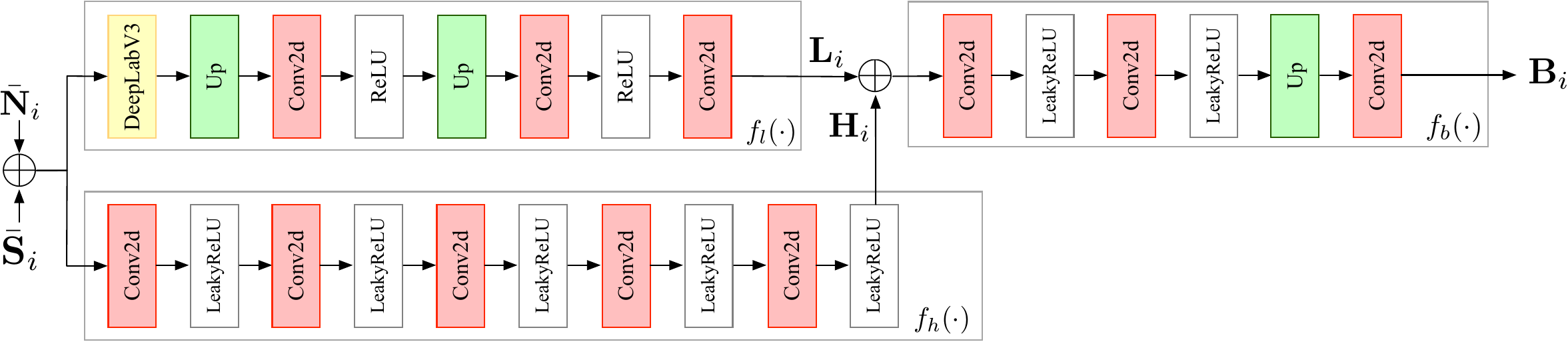}}
\vspace{-0.5cm}
    \centering\caption{\textbf{Model Architecture for multi-frame encoder $f_e(\cdot)$.} Note we stack two views together and omit the superscript $v$. The final bi-plane feature is obtained by summing the feature for each frame $\mathbf{B}_i$. $\oplus$ denotes channel-wise concatenation. }
    \label{fig:arch1}
\end{figure*}

\textbf{Multi-Frame Encoder. }We show the architecture for $f_e(\cdot)$ in Figure \ref{fig:arch1}. For simplicity of notation, we refer to all input features as these after concatenating the front and back views, \emph{e.g.} $\bar{\mathbf{N}} \in \mathbb{R}^{2 \times 512 \times 512 \times 3}$, and thus discarding the view dependency in the superscript and assume all features below have a batch size of 2. In $f_l(\cdot)$, the DeepLabV3 \cite{chen2017rethinking} backbone produce a feature map of shape $64 \times 64 \times 256$, and the output channels for the Conv2d are [128, 128, 96] respectively. All upsampling blocks are implemented as $2\times$ bilinear interpolation, thus the dimension for $\mathbf{L}_i$ is $256 \times 256 \times 96$. In $f_h(\cdot)$, the output channels for the Conv2d are [64, 96, 96, 96, 96] respectively. Note we follow \cite{trevithick2023real} to include positional encoding before the first convolution block, thus its input channel is 6 instead of 8. Except that the first convolution block in $f_h(\cdot)$ has a stride 3, all other blocks have a stride 1. The final dimension for $\mathbf{H}_i$ is the same as $\mathbf{L}_i$. In $f_b(\cdot)$ the output channels for the Conv2d are [256, 128, 96], and the biplane feature $\mathbf{B}_i$ has a shape of $512 \times 512 \times 96$.

\begin{figure*}
{\includegraphics[width=\textwidth]{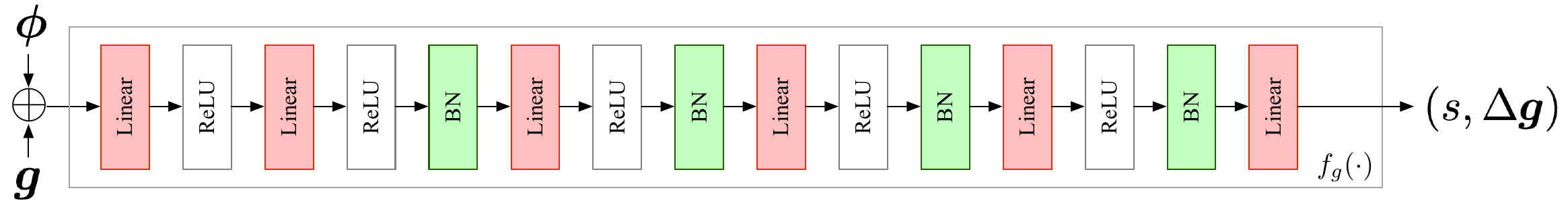}}
\vspace{-0.5cm}
    \centering\caption{\textbf{Model Architecture for canonical geometry decoder $f_g(\cdot)$.} }
    \label{fig:arch2}
\end{figure*}

\noindent
\textbf{Canonical Geometry Decoder. }We show the architecture for $f_g(\cdot)$ in Figure \ref{fig:arch2}. For each grid vertex $\bm{g}$, we sample the feature on $\mathbf{B}_i$ to obtain the feature $\bm{\phi} \in \mathbb{R}^{96}$. The output channels for each Linear block is [64, 64, 64, 64, 4]. Note here we include BatchNorm (BN) \cite{ioffe2015batch} and treat each vertex as a batch sample. We observe this module can be used to replace geometric initialization \cite{yariv2021volume} to ensure a valid mesh at initial steps, \emph{i.e.} avoid situations where the network predicts all positive or negative SDF values.

\noindent
\textbf{Skinning Weight Decoder. }We show the architecture for $f_s(\cdot)$ in Figure \ref{fig:arch2}. Except that the last linear layer has an output dimension of 161, all other modules follow the same architecture as $f_g(\cdot)$.

\noindent
\textbf{Pose-dependent Deformation Decoder. }We show the architecture for $f_c(\cdot)$ in Figure \ref{fig:arch2}. We first render front and back position maps $\mathbf{P}_t \in \mathbb{R}^{512 \times 512 \times 3}$ follow \cite{li2024animatable} and concatenate with the rendered front and back images of the inferred canonical mesh, and forward it to $f_d(\cdot)$ (with the same architecture as $f_e(\cdot)$) to produce a residual biplane $\hat{\mathbf{B}}_t \in \mathbb{R}^{512 \times 512 \times 96}$. We then sample pixel feature $\bm{\psi}_t \in \mathbb{R}^{96}$ as the feature for each vertex in the canonical mesh. The output channels for each Linear block is [64, 64, 64, 64, 3].

\section*{E. More Animation Comparison}
\begin{figure}[htp!]
{\includegraphics[width=0.46\textwidth]{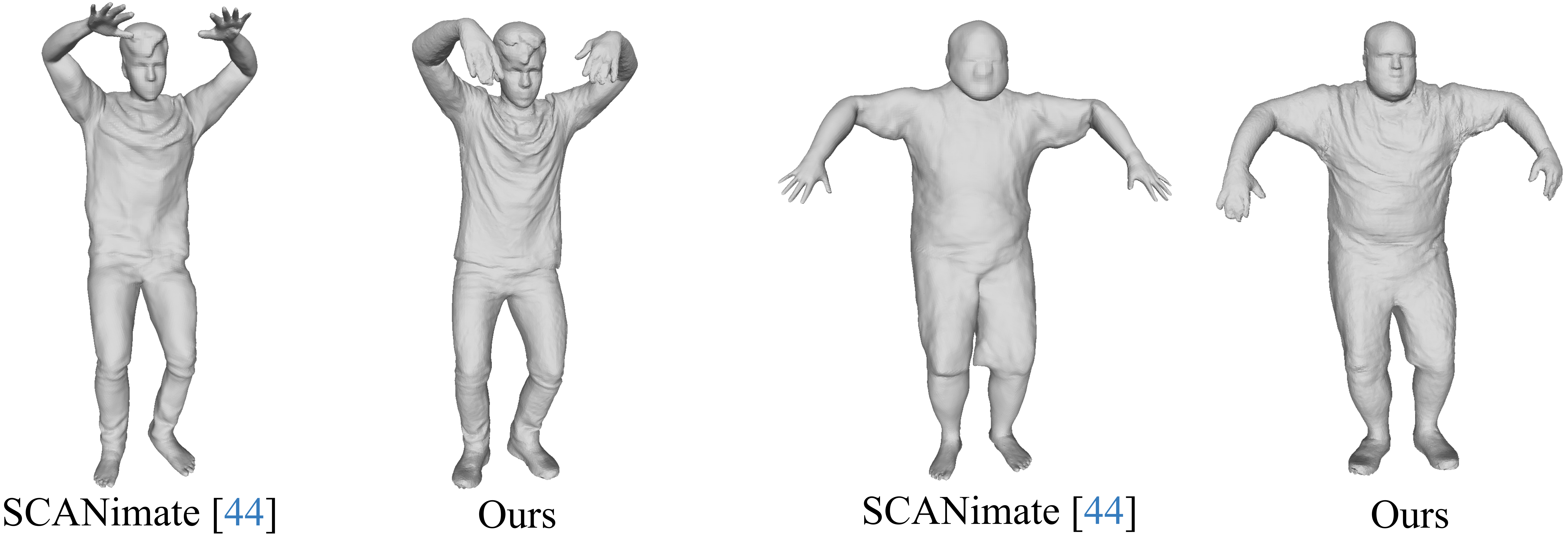}}
\vspace{-0.1cm}
\centering\caption{\textbf{Animation comparison with SCANimate.} For \cite{saito2021scanimate}, we use FRESA reconstructions as reference posed meshes. Note that hand motions are missing as it is SMPL-based. }
    \label{fig:rebuttal2}
\vspace{-0.3cm}
\end{figure}

In this section, we compare with SCANimate \cite{saito2021scanimate}, which optimizes personalized skinning weights and canonical shapes jointly in an implicit field. As shown in Figure \ref{fig:rebuttal2}, while such approach produces smooth deformation, the use of implicit field results in low geometry resolution and thus missing fine-grained details. Moreover, \cite{saito2021scanimate} rely on time-consuming per-subject fitting and 3D posed meshes as inputs, whereas our method can achieve \emph{instant} feed-forward reconstruction from \emph{few images}.

\section*{Comparison with Reconstruction Methods}
\begin{figure}[htp!]
{\includegraphics[width=0.49\textwidth]{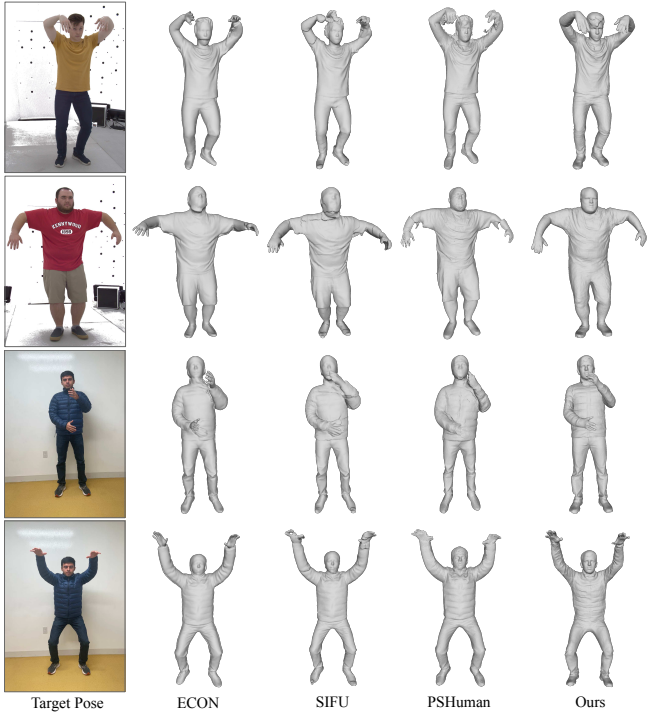}}
\vspace{-0.5cm}
\centering\caption{\textbf{Qualitative comparison with single image reconstruction methods.} Our method produces high-quality geometry details comparable to ECON \cite{xiu2023econ}, SIFU \cite{zhang2024sifu}, and PSHuman \cite{li2024pshuman} on both dome data and phone photos. }
    \label{fig:rebuttal3}
    \vspace{-0.5cm}
\end{figure}

In this work, we aim to generate personalized avatars that can be \emph{realistically animated} driven by \emph{novel poses}. In contrast, other baselines like \cite{saito2019pifu, saito2020pifuhd, xiu2022icon, xiu2023econ, zhang2024sifu, li2024pshuman} are often characterized as single-image reconstruction method, which focuses on recovering the geometry for the \emph{input pose} only, and animates posed avatars using a fixed skinning weights. Hence they do not study avatar animation and thus are \emph{not closely related to our work}. Moreover, in the experiments we evaluate the animation quality on \emph{unseen} poses and predict pose-dependent deformation to recover fine-grained details like wrinkles, thus a fair comparison is difficult to perform with the reconstruction methods. For completeness, we show in Figure \ref{fig:rebuttal3} that our method can produce high-quality geometry details comparable to \cite{xiu2023econ, zhang2024sifu, li2024pshuman}, thanks to the effective prior learned from diverse subjects. Considering fairness of evaluation, we do not quantitatively benchmarking reconstruction quality on our dataset.

\section*{F. Texture Reconstruction}
\begin{figure}[htp!]
\vspace{-0.2cm}
{\includegraphics[width=0.48\textwidth]{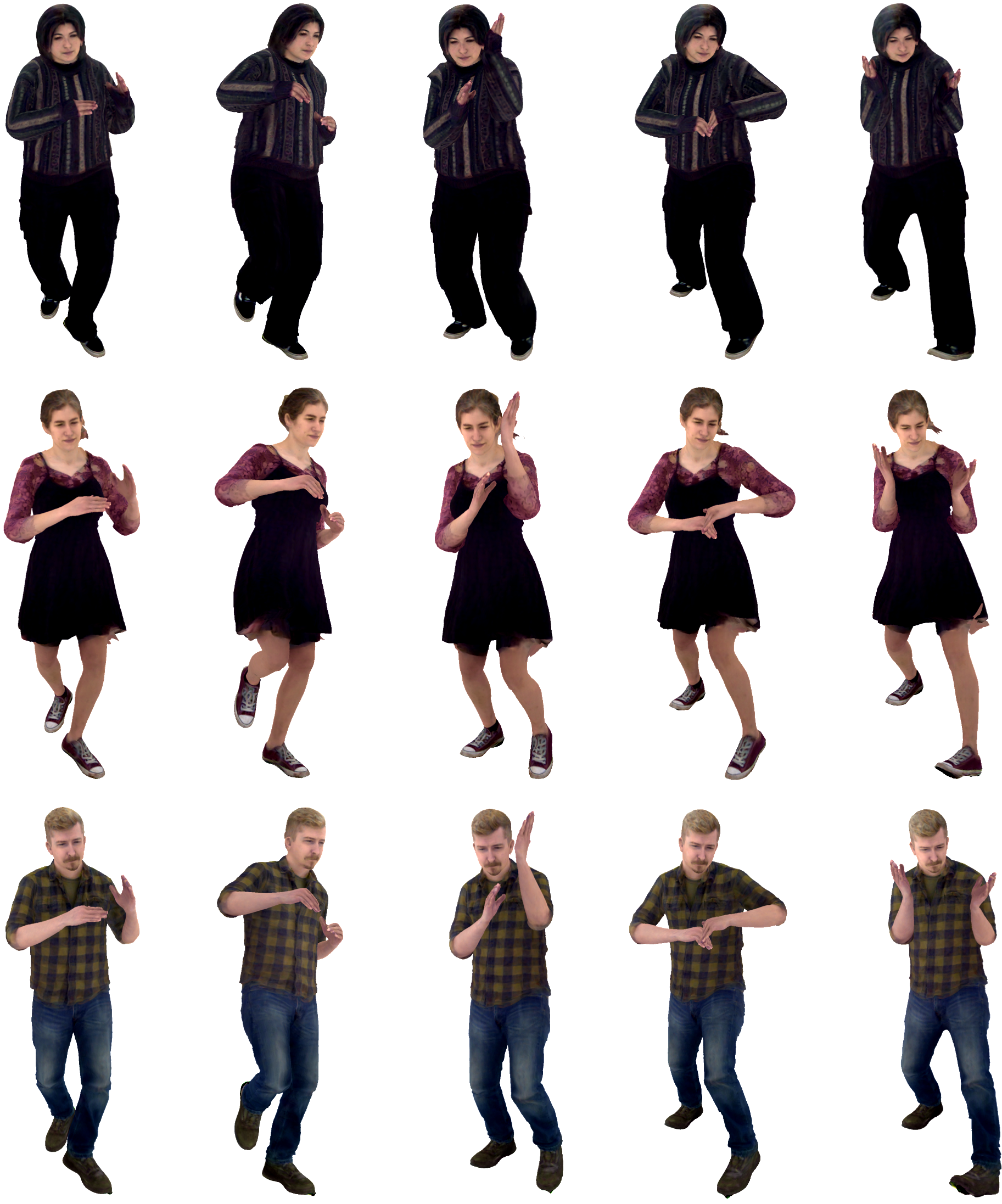}}
\vspace{-0.5cm}
    \centering\caption{\textbf{Results of Textured Meshes. }Our method can be extended to produce high-resolution texture for realistic rendering.}
    \label{fig:texture}
    \vspace{-0.3cm}
\end{figure}

Our method can be extended to generate texture for the reconstructed personalized meshes. In this section we provide one sample implementation for texture reconstruction. Specifically, we first unpose lifted surface meshes with back-projected vertex color (refer to Figure \ref{fig:ab_dbini1} as an example) and render the RGB images as input. We then encode the RGB images into a separate bi-plane feature (using a encoder with the same architecture as $f_e(\cdot)$), and pose the canonical avatar by the target pose vector $\bm{p}_t$. For each rendered pixel of the posed avatar mesh, we use the corresponding 3D position on the canonical mesh to sample the bi-plane feature, which is forwarded to a MLP decoder (with the same architecture as $f_c(\cdot)$) to predict the RGB color for that pixel. We show in Figure \ref{fig:texture} that this approach produces realistic rendering results.

\section*{G. Failure Cases}
\label{sec:texture}

\begin{figure}[htp!]
{\includegraphics[width=0.48\textwidth]{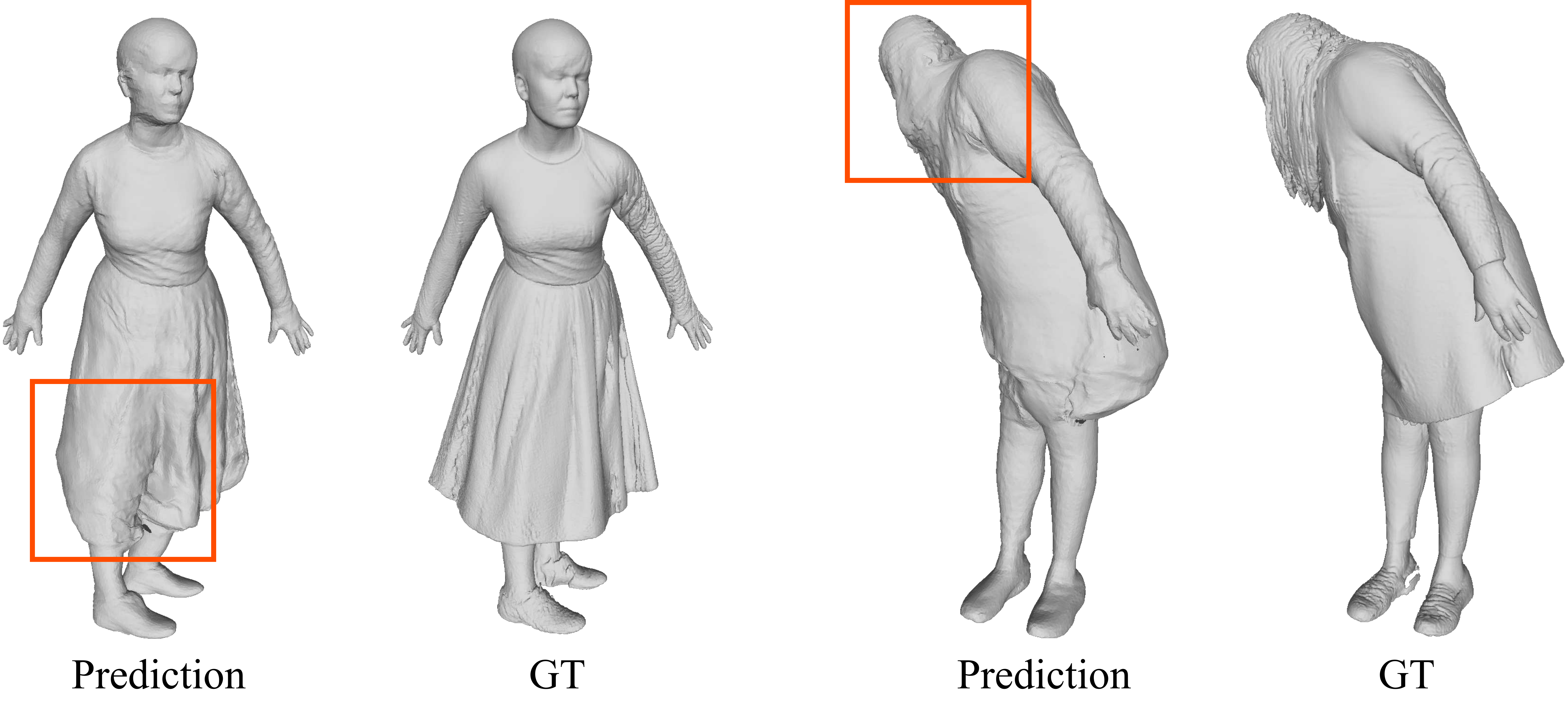}}
\vspace{-0.5cm}
    \centering\caption{\textbf{Failure Cases.} With only the pose vector as condition, our method fails to produce complex hair motions and dynamics of extremely loose garments.}
    \label{fig:fail}
\end{figure}

In our method, the deformation module is only conditioned on a skeletal pose vector, which is deficient to model complex dynamics such as motions of hair or extremely loose garments like a long dress. We show failure cases in Figure \ref{fig:fail}, where the results posed deformation do not match the real dynamics. Future works are encouraged to explore more comprehensive pose conditions or physics-inspired models to tackle this issue.  

\begin{figure*}[htp!]
{\includegraphics[width=\textwidth]{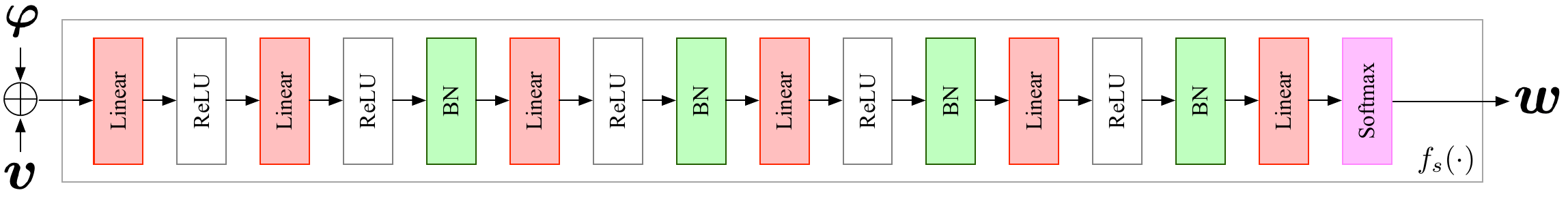}}
\vspace{-0.5cm}
    \centering\caption{\textbf{Model Architecture for skinning weight decoder $f_s(\cdot)$.} }
    \label{fig:arch3}
\end{figure*}

\begin{figure*}[htp!]
{\includegraphics[width=0.85\textwidth]{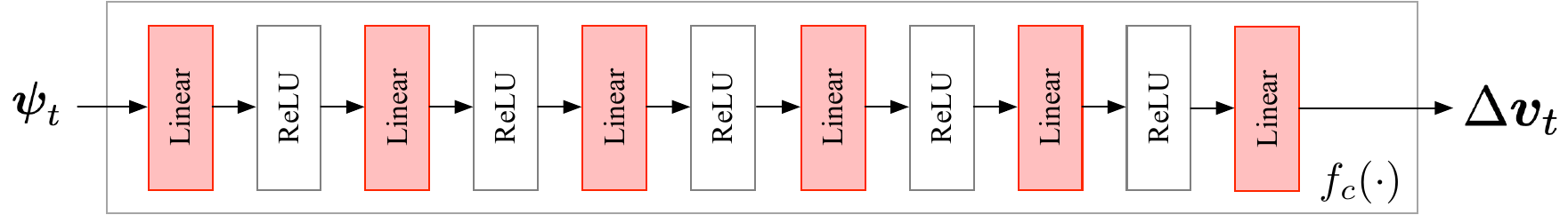}}
\vspace{-0.1cm}
    \centering\caption{\textbf{Model Architecture for pose-dependent vertex displacement decoder $f_c(\cdot)$.} }
    \label{fig:arch4}
\end{figure*}
{
    \small
    \bibliographystyle{ieeenat_fullname}
    \bibliography{main}
}

\end{document}